\documentclass{article} % For LaTeX2e
\pdfoutput=1

\usepackage{hyperref}

\usepackage{nips12submit_e,times}
\usepackage{amsfonts}
\usepackage{graphicx}
\usepackage{cite}

\usepackage{float}

\usepackage{amsmath} 
\usepackage{amssymb} 
\usepackage{amsthm}

\title{Opening the black box of deep learning}

\author{Dian Lei , Xiaoxiao Chen , Jianfei Zhao\\
School of Mechatronics Engineering and Automation,\\
        Shanghai University,
         Shanghai 200072, China }

\nipsfinalcopy % Uncomment for camera-ready version

\begin{document}

\maketitle

\begin{abstract}
The great success of deep learning shows that its technology contains profound truth, and understanding its internal mechanism not only has important implications for the development of its technology and effective application in various fields, but also provides meaningful insights into the understanding of human brain mechanism. At present, most of the theoretical research on deep learning is based on mathematics. This dissertation proposes that the neural network of deep learning is a physical system, examines deep learning from three different perspectives: microscopic, macroscopic, and physical world views, answers multiple theoretical puzzles in deep learning by using physics principles. For example, from the perspective of quantum mechanics and statistical physics, this dissertation presents the calculation methods for convolution calculation, pooling, normalization, and Restricted Boltzmann Machine, as well as the selection of cost functions, explains why deep learning must be deep, what characteristics are learned in deep learning, why Convolutional Neural Networks do not have to be trained layer by layer, and the limitations of deep learning, etc., and proposes the theoretical direction and basis for the further development of deep learning now and in the future. The brilliance of physics flashes in deep learning, we try to establish the deep learning technology based on the scientific theory of physics.
\end{abstract}

\maketitle

\section{Introduction}\label{1}

Deep learning is the main representative of the breakthrough in artificial intelligence today, it has reached nearly human level in image classification \cite{1}, speech recognition \cite{2}, natural language processing \cite{3} and so on. The method of deep learning is developing rapidly, which almost subverts all branches of computer vision field. However, the fundamental problem of deep learning at present is the lack of theoretical research on its internal principles, and there is no accepted theoretical explanation, namely, the so-called black box problem: Why use such a deep model in deep learning? Why is deep learning successful? What's the key inside? The lack of theoretical basis has led to the academic community being unable to explain the fundamental reason for the success of deep learning. The theoretical basis is not clear, and we simply do not know from what angle to look at it. The black box model is purely based on data without considering the physical laws of the model, it lacks the ability to adhere to mechanistic understandings of the underlying physical processes. Hence, even if the model achieves high accuracy but it lacks of theoretical support, it cannot be used as a basis for subsequent scientific developments \cite{4}. We must not rely solely on intuition designed algorithmic structures and several empirically tried examples to prove the general validity of an algorithm. This research method has the potential to learn false modes from non-generic representations of data, the explanatory nature of the model is very low, and the resulting research results are difficult to pass on in the long term. As people's new ideas have been replaced by more and more complex model architectures, which are almost invisible under the weight of layers of models, calls for attention to the explanatory nature of machine learning are also getting higher. Therefore, we need to thoroughly understand the entire system operation of deep learning. We need to explain what the most fundamental problems are in the field of deep learning and whether these fundamental issues are mature enough to be accurately described in mathematical and physical languages.

The great success of deep learning shows that its technology contains profound truth, but the most widely understood way is mathematical analysis, so far, very little attention has been paid to its scientific issues. However, purely mathematical explanations may lead to misdirection. For example, the neural network is mathematically trying to approximate any function. In mathematics, it has been proved that a single-layer neural network can approximate any function if it is long enough, this viewpoint has greatly hindered the development of neural networks, this is why most people in the past neglected multi-layer networks for a long time and without studying in depth. Only a small number of people such as Yann LeCun, Geoffrey Hinton, and Yoshua Bengio still insist on research in multi-layer neural networks \cite{5}. Therefore, from the great successes achieved in deep learning, it is far from enough to explain deep learning in mathematics, and the technology of deep learning needs to be based on scientific theory.

As deep learning has made breakthroughs in many aspects such as images, phonetics, and text, methods based on deep learning are increasingly being applied in various other fields, for example, recently effective in solving many-body quantum physics problems has also been proved. Therefore, the theory of deep learning methods must reflect some objective laws of the real world. obviously the most basic and universal theory is quantum physics and statistical physics. What is science? Physics is the most perfect science that has been developed so far. Just as most engineering disciplines are based on physics, the engineering foundation for deep learning now and in the future will be physics. We need to describe the deep learning concept model in the language of physics, so that we can scientifically guide the development and design of deep learning. From this we say that the key to the current and future success of artificial intelligence depends not only on the mathematical calculation, but also on the laws of physics. The theory of deep learning requires physics.

The data in the information world is divided into two different types of data: one is symbolic data, which is designated by our humans; the other is physical data, which objectively reflects the data of the real world, any actual data set we care about (whether it is a natural image or a voice signal, etc.) is physical data. Reference \cite{6} shows that the reason why neural networks can perform classification tasks well is that these physical data x follow some simple physical laws and models can be generated with relatively few free parameters: for example, they may exhibit symmetry, locality, or a simple form as an exponent of a low-order polynomial; and symbolic data, such as "variable y=cat" is specified by humans, in this case the symmetry or polynomial is meaningless, and they are not related to physics. However, the probability distribution of non-physical data y can be obtained by Bayes' theorem using the physical characteristics of x. In the reference \cite{4}, a Physics-guided Neural Networks (PGNN) is proposed, which combines the scientific knowledge of physics-based models with the deep learning. The PGNN leverages the output of physics-based model simulations along with observational features to generate predictions using a neural network architecture. Reference \cite{7} shows that deep learning is intimately related to one of the most important and successful techniques in theoretical physics, the renormalization group (RG). Reference \cite{8} using DBM and RBM to represent quantum many-body states illustrates why the depth of neural networks in the quantum world is so important, revealing the close relationship between deep neural networks and quantum many-body problems. Reference \cite{9} establishes a mapping of tensor network (TN) based on quantum mechanics and neural network in deep learning. Reference \cite{10} mentions that people have found more and more connections between basic physics and artificial intelligence, such as Restricted Boltzmann Machine and spin systems, deep neural networks and renormalization groups; the effectiveness of machine learning allows people to think about the deeper connection between physics and machine learning, and perhaps it can help us gain insights into intelligence and the nature of the universe.

The research of the above reference mainly takes the neural network as a computational tool, or as a method to solve the quantum many-body problem. This dissertation studies the artificial deep neural network as a real physical system, considers that the neural network model is a real physical model. The goal of deep learning training is to obtain the neural network system model which accords with the physical laws by the interaction or response between the neural network system and the input physical information. Because the deep neural network is a physical system, its trained model and its evolution in training must meet the laws of physics.

This dissertation analyzes the principles of physics embodied in deep learning from three different perspectives: microscopic, macroscopic, and world view, and describes deep learning with physics language, aiming to provide theoretical guidance and basis for further study and future development direction, and tries to establish the technology of deep learning based on the scientific theory of physics.

\section{A microscopic view of deep learning}\label{2}

The biggest rule of the universe is that the world is made up of microscopic particles such as atoms, electrons and photons, which obey quantum mechanics. Quantum mechanics is the science of studying the motion law of the microscopic particles in the material world, so the neural network model of deep learning as a physical system requires that the model must be governed by quantum mechanics. The following explains deep learning from the basic principles of quantum mechanics.

The human brain neural network is composed of atoms, the number of billions of neurons is the same, and the computational methods of the human brain should be similar. The neural network, as an interactive quantum many-body system, determines the deep learning system to be described by the wave function. The coordinate operator, momentum operator (corresponding translation operator), angular momentum operator (corresponding rotation operator), energy operator, and spin operator in the neural network are the most basic and important physical quantity or mechanical quantity operator.

\subsection{The physical meaning of neurons}\label{2.1}

Information has both physical and symbolic meanings, so neurons also have two meanings: 1) physical, 2) symbolic mappings. Now discuss the meaning of its physics. In this dissertation, the first hypothesis is that the neuron is the scattering source of the quasi-particle wave and the superposition of receiving the quasi-particle wave. First look at a classic physics experiment---Young's double slit experiment.

\begin{figure}[H]
\centering
    \includegraphics[width=3in,height=2in]{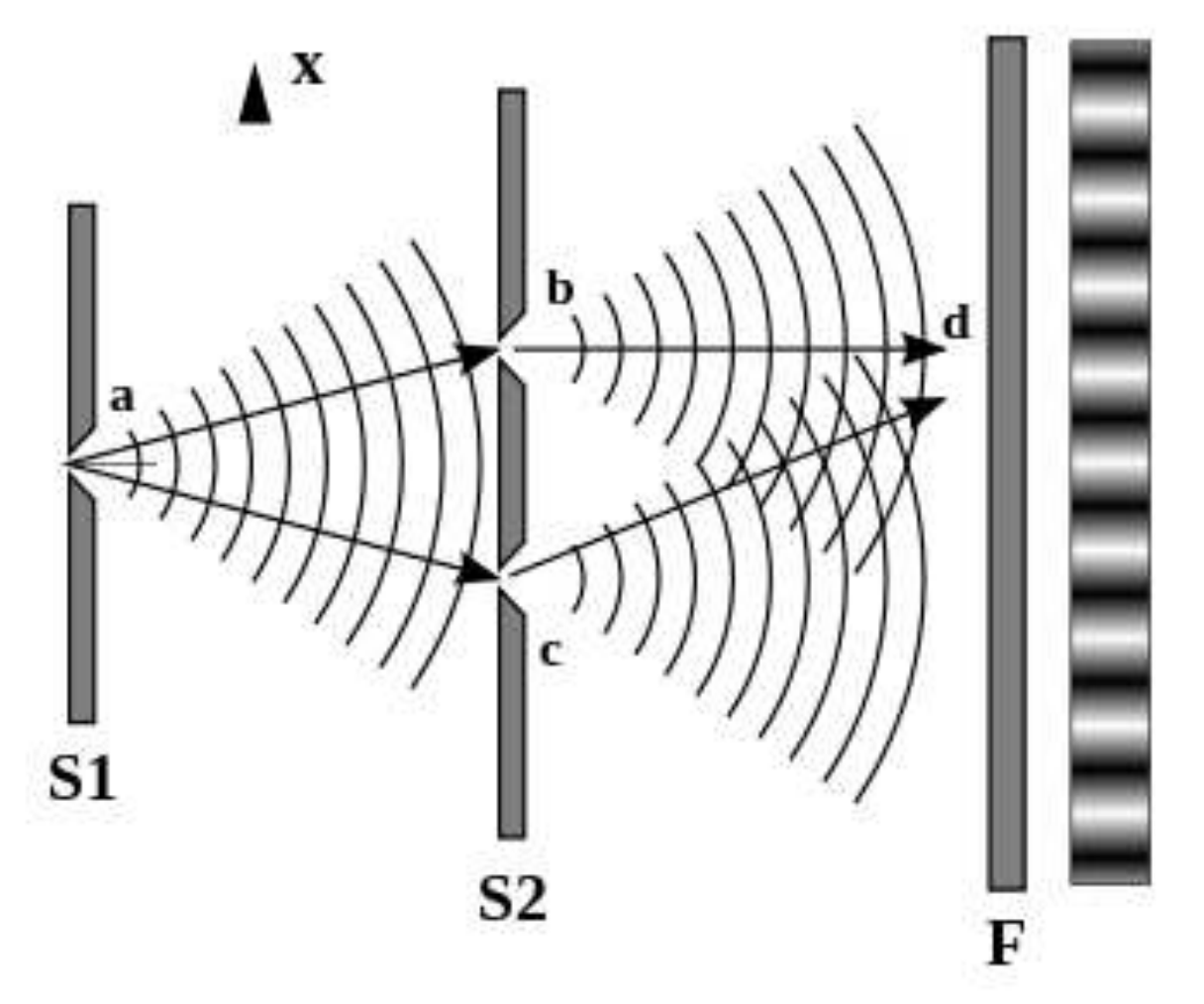}
    \caption{Young's double slit experiment.}
    \label{figure1}
\end{figure}

As shown in Figure \ref{figure1}, the electrons are diffracted through the aperture a, and then diffracted and interfered by b and c. The bright diffraction fringes and patches at F indicate that there is a greater probability of electrons appearing there, and the dark part is there is little or no chance of the appearance of electrons. This dissertation holds that neurons act as electron diffraction interference. When we look at a neuron as a physical unit, the neuron is a scattering potential well that causes scattering of quasiparticles (perhaps this scattering originates from the quantum effect in the microtubules of the neurons, perhaps the electron-phonon coherence coupling in the biological system, perhaps some other kind of elementary excitation). Therefore, the output after the input of the neuron calculation is like the scattering output of the electrons through the circular hole, and the law is determined by the quantum physics theorem. The physical meaning of neurons indicates that, as white light can be scattered as red, orange, yellow, green, cyan, blue, violet, it is a natural classifier, calculator.

In the Young's double slit experiment, whatever the input is light, or electrons, polarized electrons, neutrons, and any particle including atomic and subatomic levels will cause diffraction and interference effects. In the same way, the physical meaning of neurons indicates that neurons are inherently capable of discriminating characterizations and are inherently capable of excellent generalization.

The structure of Young's double slit experiment provides the foundation for a universal quantum neural network. The physical model of neurons assumed in this dissertation is:

The input of neurons is a multichannel wave function, for example, the input image is a wave function of multiple pixel points, and the photon (quasi-particle) wave function is superimposed to become the new scattering source output. The state value of the neuron is the number of quasi-particles (probability) of excitation after superposition, if you visualize multiple or large numbers of neurons, you see images of the same nature as the interference diffraction experiment¡ªstreaks and patches.

\begin{figure}[H]
\centering
\includegraphics[width=3.3in,height=2in]{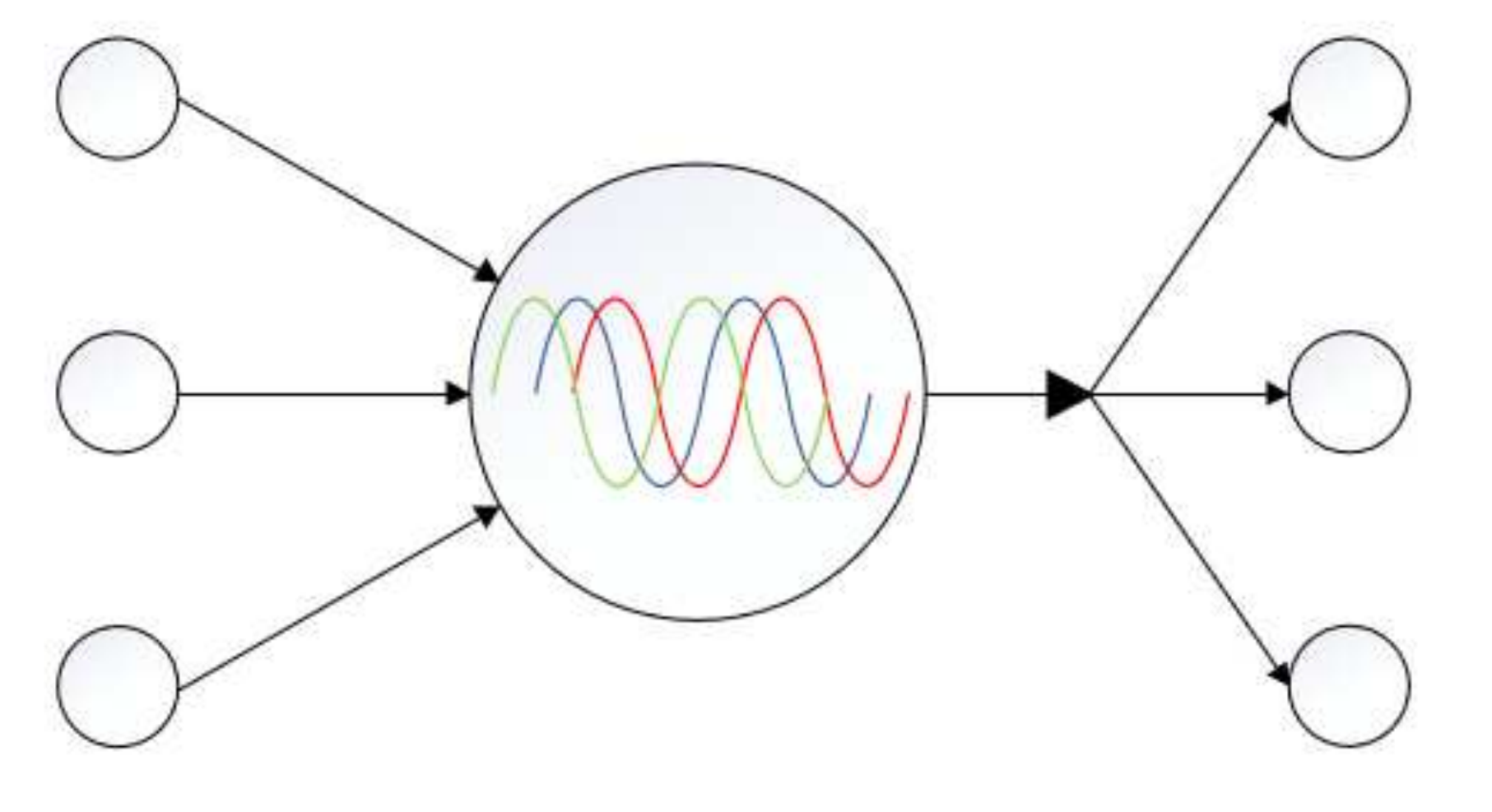}
\caption{Neuron physical model diagram.}
\label{figure2}
\end{figure}

The fundamental of this model is that the neuron is a physical model, which reflects the probability of its state value, and the related theories and deep learning techniques discussed later in this dissertation can confirm the correctness of its hypothesis.

\subsection{Quantum physical model of CNN}\label{2.2}

In this section, we first establish the quantum physics model of the Convolutional Neural Network, and then make a scientific analysis of the CNN based on this model, give a physical explanation of the success of the convolutional neural network, and explore the prospects for further development.

The structure of Young's double slit experiment provides the foundation for a universal quantum neural network, as shown in Figure 3:
\begin{figure}[H]
\centering
\includegraphics[width=4in,height=2in]{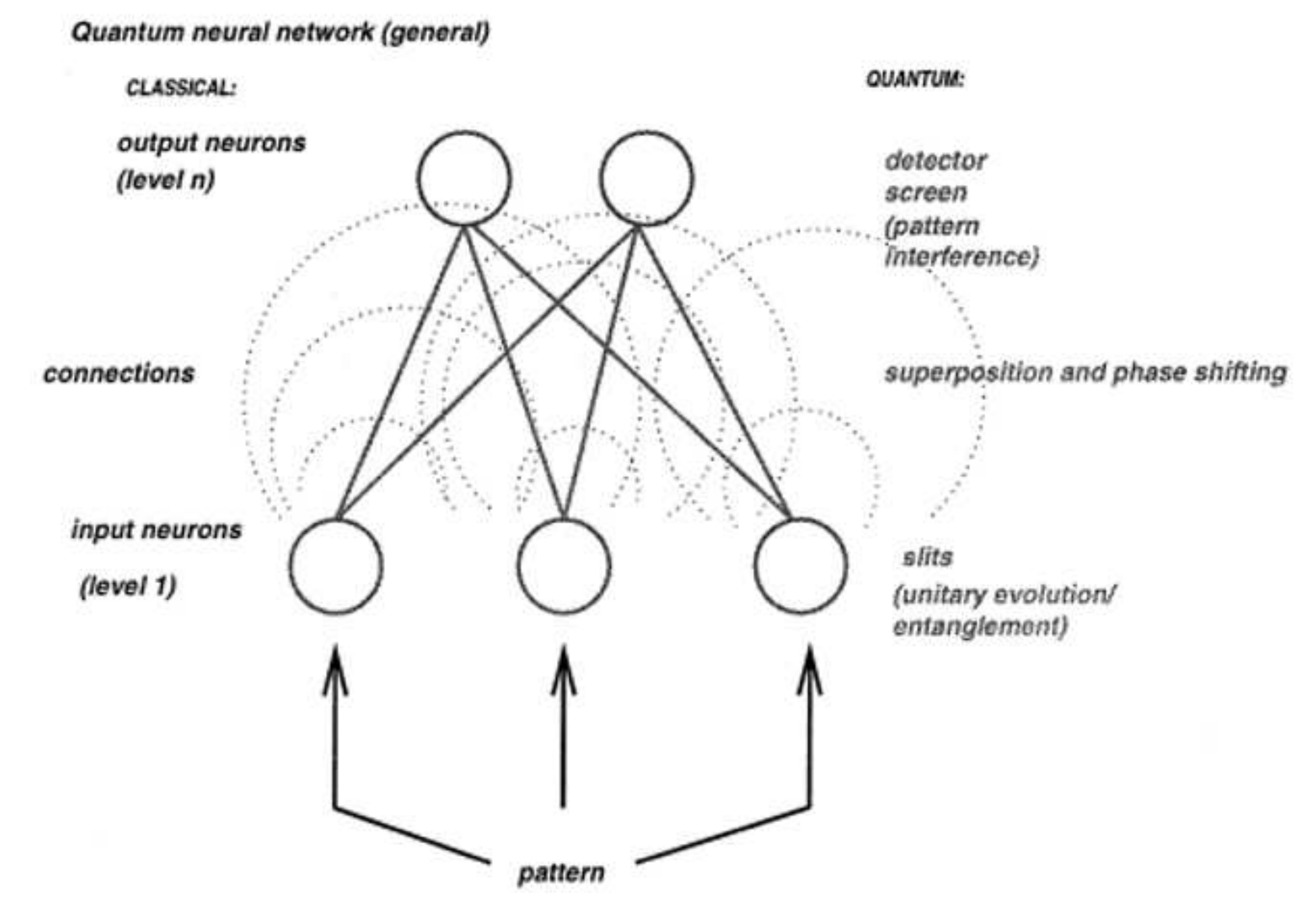}
\caption{Universal quantum neural network.Figure courtesy:\cite{11}}
\label{figure3}
\end{figure}

Similar to the quantum neural network in reference\cite{11}, this dissertation considers that the neural network model of deep learning is the physical model of the interference diffraction of photon or quasiparticle. The difference is that the input of the network is changed from the photon gun to the image of the input layer. The detection screen is made up of many neurons. Some neurons superimpose the input photon or quasi-particle states on each path to obtain intensity (probability of photons or quasiparticles) and re-scattering output, the intensity values of all neurons constitute an interference diffraction pattern. In addition to the input, the interference diffraction pattern is also closely related to the structure of the neural network; the connecting line between neurons is related to the excitation mode or interaction potential of the neuron and the structure of the neural network, and has nothing to do with the input, which is consistent with the usual concept of artificial neural network.

Is this model correct? Let's look at two figures:
\begin{figure}[H]
\centering
\includegraphics[width=3in]{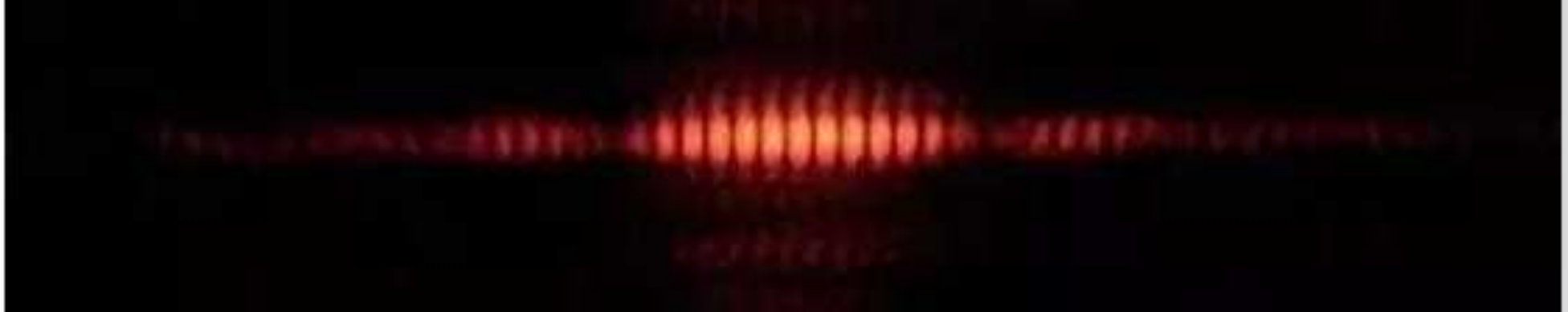}
\caption{ A real interference pattern}
\label{figure4}
\end{figure}

\begin{figure}[H]
\centering
\includegraphics[width=3in]{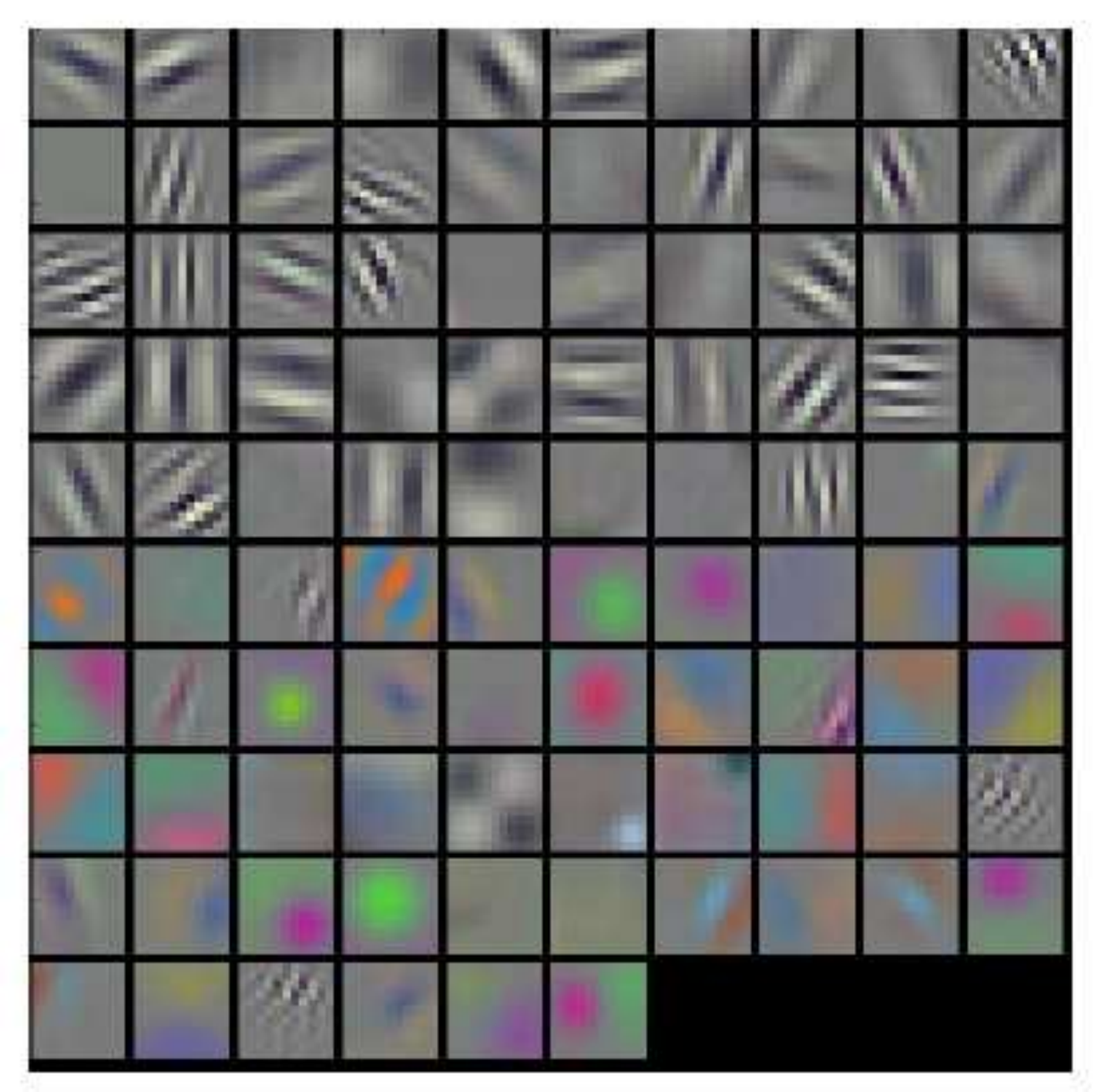}
\caption{ The first convolution layer of a typical convolution neural network after training.Figure courtesy:http://cs231n.github.io/understanding-cnn/}
\label{figure5}
\end{figure}

 Figure \ref{figure4} is a diffraction interference fringe obtained from a real double-slit experiment, which contains the total intensity distribution information of double-slit interference and single-slit diffraction. Figure \ref{figure5} shows the first convolution layer after a typical CNN training, it also sees stripes and patches. Comparing the two graphs, we can see that the two are very similar, and all of them are stripes and patches, which fully shows that our neural network model is the quasi-particle diffraction interference model in physics. The neuron receives multiple signals from the previous layer, performs quantum superposition, if the interference superposition elimination, then the neuron remains silent; if the interference superposition intensity is extremely high (the stripe or patches in the figure), then the neuron activates the impulse, emits the superposition intensity output signal. This also explains why physiological neurons may remain silent even if they have strong signal input.

From the first layer of the convolutional layer in Figure \ref{figure5}, we can see that most people think that the first layer of convolution is to learn the edge of the image, and what the later convolution layer learned is no one can understand. In this theoretical model, the first and the back layers are interference diffraction fringes! Interferometric diffraction or scattering experiments in physics can analyze and discover elements, and deep learning neural networks can be used for a wide range of applications such as image recognition because its mechanism is also the mechanism of interference diffraction or scattering experiments. Each fringe or patch reflects the momentum and angular momentum characteristics of the system, and the momentum or wave vector mechanics respectively represent the translation and rotation invariance of the space.

Let us first look at the relationship between translation operation and convolutional neural networks.

\subsubsection{CNN and translation operation}\label{2.2.1}

According to quantum mechanics, the space translation operator is:
\begin{equation}
    \hat D(a) = {e^{ia.\hat p/h}}  \text{or}  \hat D(a) = {e^{i\hat k.a}}\label{equation1}
\end{equation}

Where $a$ is the translational magnitude, the wave vector operator $\hat k =  - i\nabla $ , the momentum operator $\hat p =  - ih\nabla $ , $\hat p \equiv h\hat k$, $h$ is the Planck constant, the wave vector $k$ and the momentum $p$ are independent of $a$, and are mechanical quantities only related to the system.

The physical quantities that describe the spatial invariance or spatial symmetry of the arbitrary function $f(x + a) = f(x)$ are the momentum or wave vector, i.e. the physical conservation quantities corresponding to the spatial symmetry are the wave vector $k$ or the momentum $p$. For example, the free particle or plane wave function $\psi (x) = C{e^{ikx - i\omega t}}$ , where $k$ is the wave vector, equal to the reciprocal of the wavelength, it describes its spatial properties. The edge boundary, texture, and color change of the image must destroy the translation symmetry. The mechanical quantity wave vectors can be used to express whether the edge, texture, color of the image is equal or unequal in space. The wave vector is the basic physical quantity quantifying these attributes.

The translation in a one-dimensional case is easy to prove:
\begin{equation}
{e^{a\frac{d}{{dx}}}}f(x) = f(x) + \sum\limits_{n = 1}^\infty  {\frac{{{a^n}}}{{n!}}} \frac{{{d^n}}}{{d{x^n}}}f(x) = f(x + a)\label{equation2}
\end{equation}

which is:
\begin{equation}
\hat D(a)f(x) = f(x + a)\label{equation3}
\end{equation}

The convolution calculation in the CNN actually reflect the translation operation, and the one-dimensional convolution in book \cite{12} for image $I(x)$ and kernel function $K(a)$ is as follows:
\begin{equation}
\begin{split}
 S(x) &= \sum\nolimits_a {K(a)I(x - a)}  \\
 &= \sum\nolimits_a {K(a){e^{ - a\frac{d}{{dx}} + a\frac{d}{{dy}}}}I(x)}  \\
 &= \sum\nolimits_a {K(a){e^{ - \hat k.a}}I(x)}  \\
 &= \sum\nolimits_a {K(a)\hat D(a)I(x)}
\end{split}\label{equation4}
\end{equation}

The two-dimensional form is as follows:
\begin{equation}
S(i,j) = \sum\nolimits_m {\sum\nolimits_n {K(m,n)I(i - m,j - n)} }\label{equation5}
\end{equation}

The entire feature map layer image is obtained by translation. To understand the image, consider the one-dimensional case, the minimum unit of translation is a pixel, if $K(a) = \left\{ \begin{array}{l}
 1,(a < r) \\
 0,(a \ge r) \\
 \end{array} \right.$ is set. For the input layer is the image, only two points are taken within the $\left| r \right|$ range, the range is small, and the output of the convolutional layer is:
\begin{equation}
S(x) = \sum\nolimits_a {K(a)\hat D(a)I(x)}  = (\hat D({\rm{0}}){\rm{ + }}\hat D({\rm{1}}))I(x) \approx (2 + \frac{d}{{dx}})I(x)
\label{equation6}
\end{equation}

So the first convolution layer sees the edge of the image element in the background, but it is not equal to the edge of the image element because of the convolution kernel.

If the image $I(x)$ has spatial symmetry with wavelength $\lambda $, or if the wave vector has spatial symmetry of $k = 2\pi /\lambda $ , i.e. $I(x) = \sin (kx)$ , then the output of the convolutional layer is:
\begin{equation}
S(x) = \int_0^r {\hat D(a)} I(x)da = \int_0^r {\sin k(x + a)} da = \frac{2}{k}\sin \frac{{kr}}{2}\sin (kx + \frac{{kr}}{2})\label{equation7}
\end{equation}

After convolution, the value still has the same wave vector $k$ volatility, but the intensity is $\frac{4}{{{k^2}}}{\sin ^2}\frac{{kr}}{2}$, the interference diffraction envelope appears, the maximum and the minimum intensity values appear in:
\[kr = n\pi \left\{ \begin{array}{l}
 n = 1,3,5, \cdots {\rm{ the \ intensity \ is \ extremely \ high}} \\
 n = {\rm{2}},{\rm{4}},{\rm{6}}, \cdots {\rm{ the \ intensity \ is \ zero}} \\
 \end{array} \right.\]

The above results show that an interferometric diffraction image is obtained by spatial convolution. When the input wavelength is longer, the intensity at the maximum by the convolution is amplified to $4/{k^2}$ times the original input wave intensity. When the convolution kernel structure is certain, only the wave vectors satisfying certain conditions will appear extremely large. Assuming that only the ¡°brightest¡± point is taken, then only the input wave vector satisfies $k = \pi /\lambda $ sine wave through the convolution after the "bright spot." That is to say, in the case of a convolutional kernel structure, the "brightest" reflects the wave vector $k$ of the sine wave. The actual wave is the superposition of sine wave, and the "brightest" point of the interferometric diffraction image will not have only one point. The interference diffraction image reflects the spatial symmetry of the input wave.

The next section explains what the physical meaning of the convolution formula (Equation \ref{equation4}) is, explains the nature of the convolution kernel and explains why the convolution calculation is not summed over the entire image range, but instead use a finite window convolution calculation such as 3*3? Why do these moving convolution windows use the same convolution kernel, the so-called shared weight? Why does a convolutional layer use multiple feature mapping layers? Why does each feature mapping layer convolutional kernels differently? Why pooling? Why use Relu activate functions, etc. In our model, the neuron is a unit that receives multiple inputs and quantum superimposes and then scatters, so a perfect answer can be obtained by using the scattering theory of quantum mechanics.

\subsubsection{Physical meaning of convolution calculations and convolution kernels}\label{2.2.2}

A general CNN interprets a convolution kernel as a filter, but what does it filter? What does it keep? It is not clear, so this argument is difficult to convince, and the filter's opinion is not clear. Convolutional neural networks have achieved great success in the field of computer vision, one of the most important components is the convolution kernel, which is primarily responsible for capturing most of the abstraction of the network. In contrast, this component is the least understood processing block because it requires the maximum computational learning \cite{13}.

This section builds the convolutional calculations on the quantum physics model described in Section \ref{2.1}. The physical basis of the interference diffraction experiment is quantum mechanical scattering theory. According to the scattering theory, the system Hamiltonian does not contain time, and the incident wave is directed at the target. The wave function $\psi$ at the position $r$ is: \cite{14}.

\begin{equation}
{\psi _k}(r) = {e^{ik \cdot r}} - \int {G(r - r')U(} r')\psi (r')dr'\label{equation8}
\end{equation}

Equation \ref{equation8} is the integral equation, the integral is performed in the entire space, the first is the input wave, and the second is the scattering wave. The physical meaning of the second item is very clear, $U$ is the interaction potential function. The input wave scatters in the $dr'$ range near $r'$ point to form a scattering point source with intensity $U(r')\psi (r')dr'$ . This point source propagates to the observation screen $r$ point according to the outgoing Green function, and the scattering of the points is superimposed as the scattering probability amplitude ${\psi _{sc}}$ .

The Green function is:
\begin{equation}
G(r,r') =  - \frac{{{e^{ikr}}}}{{4\pi \left| {r - r'} \right|}}\label{equation9}
\end{equation}

The incident wave requirement in Equation \ref{equation8} is monochromatic and planar wave, and does not do any other approximation. The equation is applicable both to elastic scattering and inelastic scattering or collision scattering. So the total intensity of a neuron scattering from a quasi particle of a certain wavelength is
\begin{equation}
S = \int {{{\left| {\psi (r)} \right|}^2}dr}\label{equation10}
\end{equation}

The integral $r$ is in the range of neuron size. According to this model, the total intensity of a neuron is defined as the probability density ${\psi ^{\rm{2}}}$ of the excited particle, so
\begin{equation}
S \approx {\left| {\psi ({\rm{0}})} \right|^2}\label{equation11}
\end{equation}

Let $S$ be the square of the convolution $s$ , that is $S=s^2$ , then the convolution $s$ is:
\begin{equation}
s = \left| {\int {G(r')U(r')\psi (r')dr}  + b} \right|
\label{equation12}
\end{equation}

Note that the left side of Equation \ref{equation8} is a wave function, and there are also wave functions in the integral term, so it is difficult to solve. As an approximate calculation, let the $\psi $ in integral as input image wave, converted to discrete expression. The volume element $dxdydz = 1$
, so the convolution for a wave vector $k$ is ${s_k}$ :
\begin{equation}
{s_k} = \left| {\sum\nolimits_{r'} {G(r';k)} U(r')\psi (r') + b} \right|
\label{equation13}
\end{equation}

Where the sum range of $r'$ is within the neuron volume range. The convolution kernel is:
\begin{equation}
K = G(r';k)U(r')\label{equation14}
\end{equation}

$\psi (r')$ is the input image wave, it is only related to $x',y'$ , and has nothing to do with other coordinates:
\begin{equation}
\begin{split}
 {s_k}& = \left| {\sum\nolimits_{x'y'z'} {G(x',y',z';k)} U(x',y',z')\psi (x',y') + b} \right| \\
  &= \left| {\sum\nolimits_{x'y'} {\sum\nolimits_{z'} {G(x',y',z';k)U(x',y',z')} \psi (x',y')}  + b} \right|
\end{split}\label{equation15}
\end{equation}

The square $s_k^2$ of $s_k$ represents the quasi-particle probability density that is excited when the incident wave (or quasi-particle) has a wave vector k (size, direction, and polarization direction). So the convolution kernel:
\begin{equation}
{K_{x'y'}}(k) = \sum\nolimits_{z'} {G(x',y',z';k)U(x',y',z')} \label{equation16}
\end{equation}

 $x',y'$ is a two-dimensional image coordinate, $s_k$ represents the convolution of the feature mapping layer, and different wave vectors have different $s_k$ (if the quasi-particles are electrons also have a direction: spin)
\begin{equation}
{s_k} = \sum\nolimits_{x'y'} {{K_{x'y'}}(k)} {\psi _{x'y'}} + b \label{equation17}
\end{equation}

The convolution of all k is a set $S = \{ {s_k}\} $.

The following is a discussion of its physical meaning and its implications for the calculation of convolutional neural networks.

\subsubsection{Quantum mechanics interpretation of CNN}\label{2.2.3}

There are various interpretations of convolutional neural networks, but according to the above quantum physics model, CNN can be perfectly interpreted. The interpretation is fundamental and deterministic, it will inspire the architectural design of deep convolutional networks.

\begin{itemize}
\item[(1)] When the inner product of the convolution is represented by the norm and the angle: $\left\| \omega  \right\|\left\| x \right\|\cos ({\theta _{(\omega ,x)}})$ , the trained inner product of the convolutional neural network can be decoupled to find the relationship between norm and angle in the feature map \cite{15}.
    \begin{figure}[H]
        \centering
        \includegraphics[width=3in,height=2in]{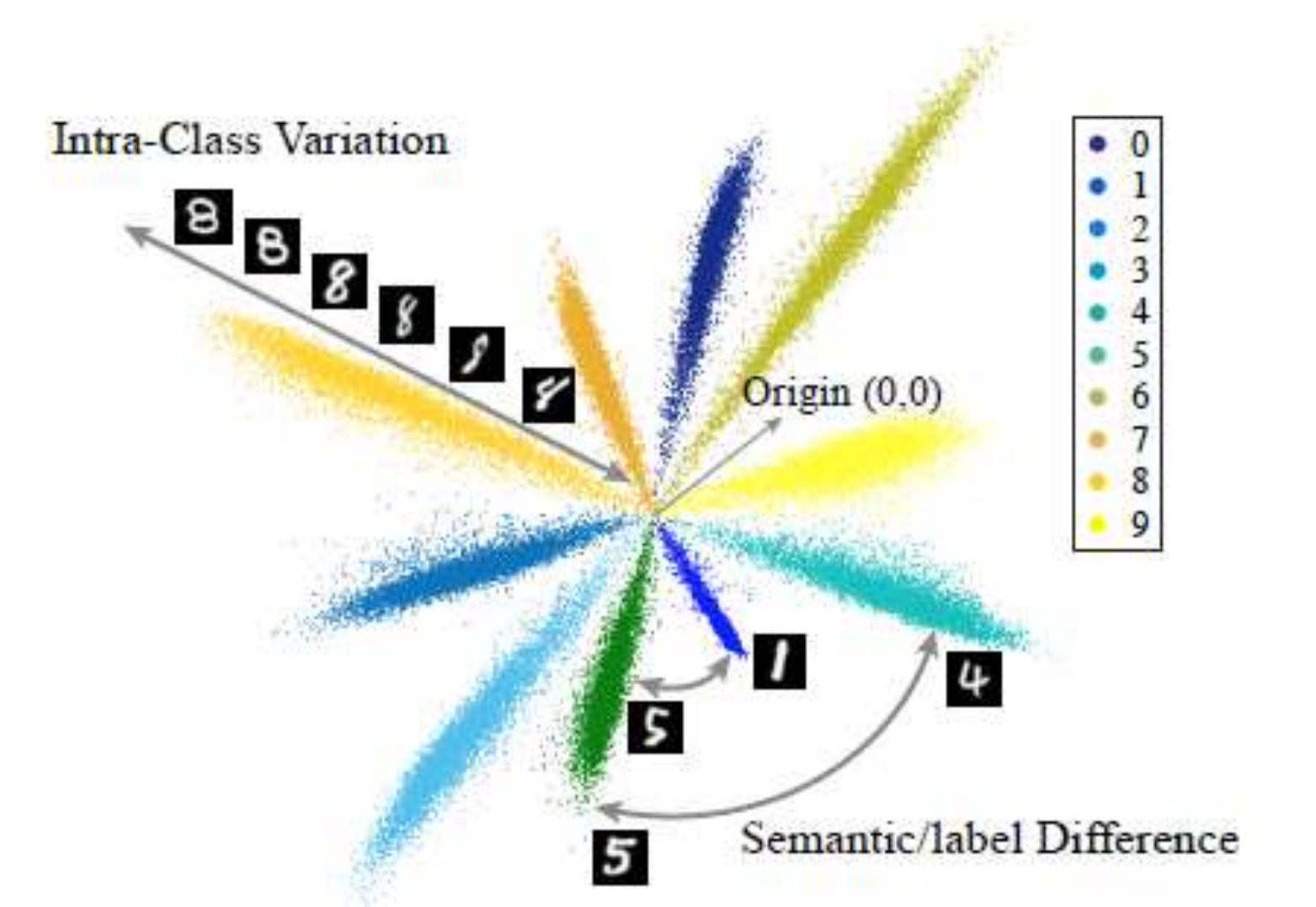}
        \caption{CNN learned features are naturally decoupled.Figure courtesy:\cite{15}}
        \label{figure6}
    \end{figure}

    If the convolution of spherical coordinates is used, the relationship between the norm and angle in the feature map is more obvious:
    \begin{figure}[H]
    \centering
        \includegraphics[width=3in,height=2in]{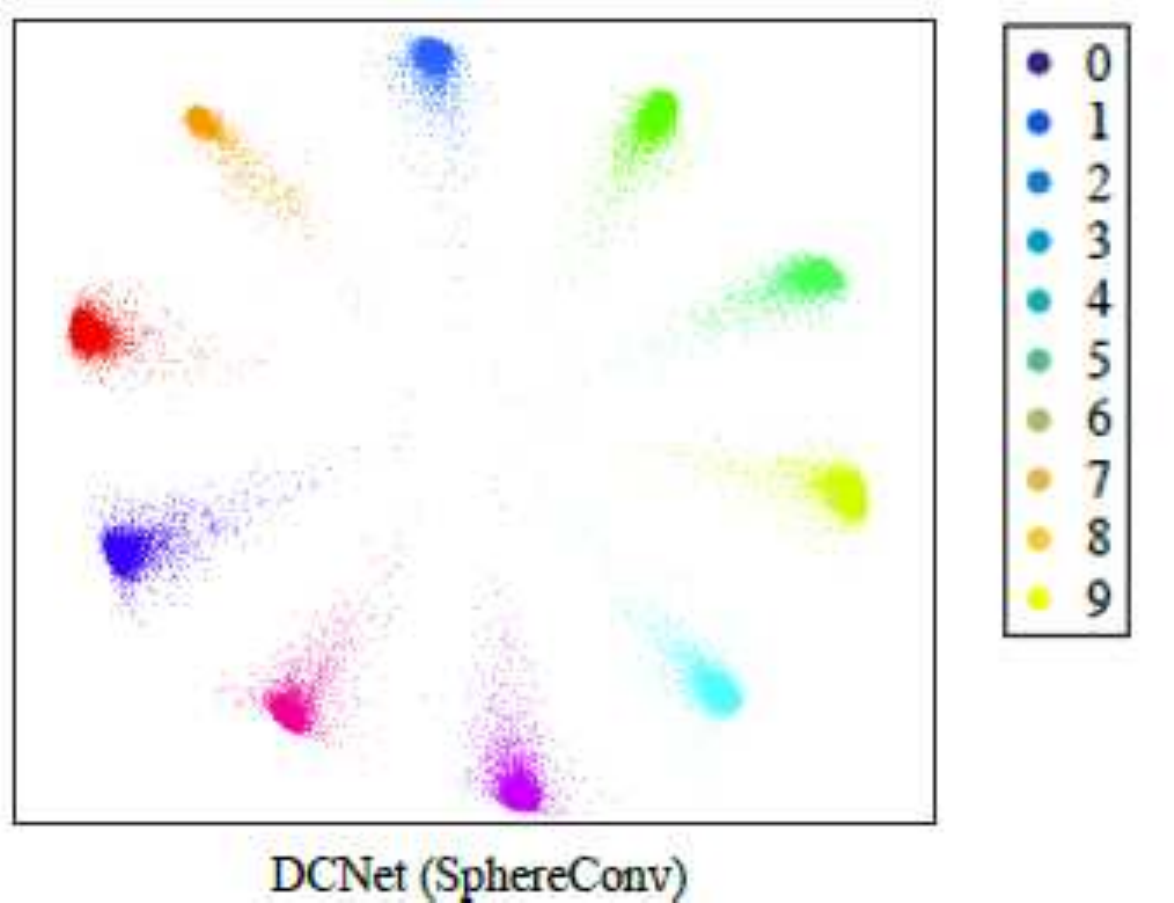}
        \caption{2D feature visualization on MNIST dataset with natural training.Figure courtesy:\cite{15}}
        \label{figure7}
    \end{figure}

    The figure shows that the angles represent semantic/label differences, and the feature norm represents within-class differences. This result can be explained in our model. It can be seen from (Equation \ref{equation14}) that the general action potential $U({\rm{r}})$ is only related to $r$ , i.e.  $U({\rm{r}}) = U(r)$ . So the convolution kernel is related to the norm and has nothing to do with the angle, the CNN feature image is related to the angle. The above experiments verify the correctness of the physical model of the convolutional neural network.

\item[(2)] The number of convolution kernels is related to the color of the image. The input requirement in the scattering equation(Equation \ref{equation8}) is a monochromatic wave, that is, the convolution kernel(Equation \ref{equation16}) is related to the wavelength of the input wave, and different wavelength should have different convolution kernel ${K_{mn}}$. Applied to the color image, convolution neural network should have different convolution kernel for different colors image. For example, they can be composed of three kinds of convolution kernels: red, green, and blue. Of course, they can also be composed of multiple kernels such as red, orange, yellow, green, blue, and purple.

\item[(3)] The number of convolution kernels is also related to the polarization direction of the incident wave. It can be seen from(Equation \ref{equation16}, \ref{equation17}) that even if the image is monochromatic, the convolution kernels of different $z$ are different. So even if the input is a black-and-white image, multiple feature mapping layers (multiple convolution kernels) are needed, just as white light contains light of all colors, and random monochromatic light also contains polarized light of all polarization directions. If the input is an electron, the number of convolution cores is related to the spin direction. In a comparable size, the number of feature map layers is much larger than the convolutional window size.

\item[(4)] The reason of partial receptive fields is not because the pixels are usually highly correlated in neighboring regions. The convolution kernel sharing of feature layer is not due to translation invariance, they are due to the limited scope of the action potential $U(r')$ of the neuron, that is, the coordinate $x',y'$ in the convolution kernel (Equation \ref{equation16}) is 0 outside the range of action potential. Therefore, the summation of the $x',y'$ in (Equation \ref{equation17}) does not require full space, i.e.the CNN should not be fully connected but partial connection. For the same feature mapping layer, all the neuron potentials $U(r')$ are the same, so the convolution kernels of all the sliding window convolutions of the same feature mapping layer are the same or share one, which explains from the scientific theory that the fundamental reason for the success of CNN is the important characteristic of CNN: partial receptive fields + weight sharing. In the actual application of CNN, the convolution kernel size (that is, the convolution window size) is often 3x3, 5x5, 9x9 and so on, scattering is mainly concentrated in the incident direction, so the window size is too large to be meaningful. However, if the 1*1 size of the convolution kernel is used, the interference effect will also be poor. In addition, considering the symmetry of $K$, convolution kernel size is generally used odd.

\item[(5)] The purpose of pooling operations is not only to reduce the size of the output eigenvectors. (Equation \ref{equation15}) is the intensity of a neuron, and the image formed by all the neurons of a convolutional feature layer is an interference diffraction pattern. Because the coherence of the wave is strong in some places and weak in some places and not even in some places, and those "bright" neurons will be the scattering sources of the subsequent convolutions, so each convolutional layer must be pooling. That is, in the convolution window area, the "bright" neurons must be selected according to the intensity values of the neurons in each convolution layer feature layer, which is also the physical reason why the biological neuron can either output or not output even if it has a signal input. Obviously, pooling is an important computing component. In the early years, many of the studies based on convolution architecture used average pooling, now they are replaced by Max Pooling. From our model we can see that in practical application, because the window size is not large, it is reasonable to select one of the most ¡°bright¡± neurons, that is, to adopt the max pooling method. It is also seen from here that the pooling is of decisive importance to the initial convolution layer, but as the number of layers increases, the interferogram sharpens and ¡°bright¡± points become less, so the effect of pooling is weakened, which is the same as the result of the reference \cite{16}. Pooling can also be explained by the renormalization group in Section\ref{3.4}.

\item[(6)] The convolution kernel of different layer is not same, because each of the "brightest" neuron positions represents the coherence of the corresponding convolution window image, which is related to the symmetry of the image and the structure of neural network. After pooling, the spacing as a new scattering source is different, so subsequent convolution kernels will also be different. That is, except for the same feature mapping layer, convolution kernels of each feature map layer of the convolution layer are different.

\item[(7)] Coherence is the most important condition for multilayer CNN. In our model, the purpose of convolution is to create the entanglement of each pixel in the image under the action of neurons, thus forming the interference diffraction fringes or patches according to the spatial translation structure. According to the quantum mechanics, this is a coherence phenomenon caused by the superposition principle of waves. The total intensity after interference superposition is not necessarily equal to the sum of the intensity of the sub-beams, may be more strong or may be equal to 0 under the interference. The important condition of coherence is the coherent wave. There are two ways to produce coherent wave, the first one is to ensure that the monochrome of the wave and the phase of each wave of the fixed, so need multi-layer convolution; the second is the interference of oneself and oneself, so need multi-layer convolution, generally need at least 3 or more.

    \begin{figure}[H]
    \centering
        \includegraphics[width=3in]{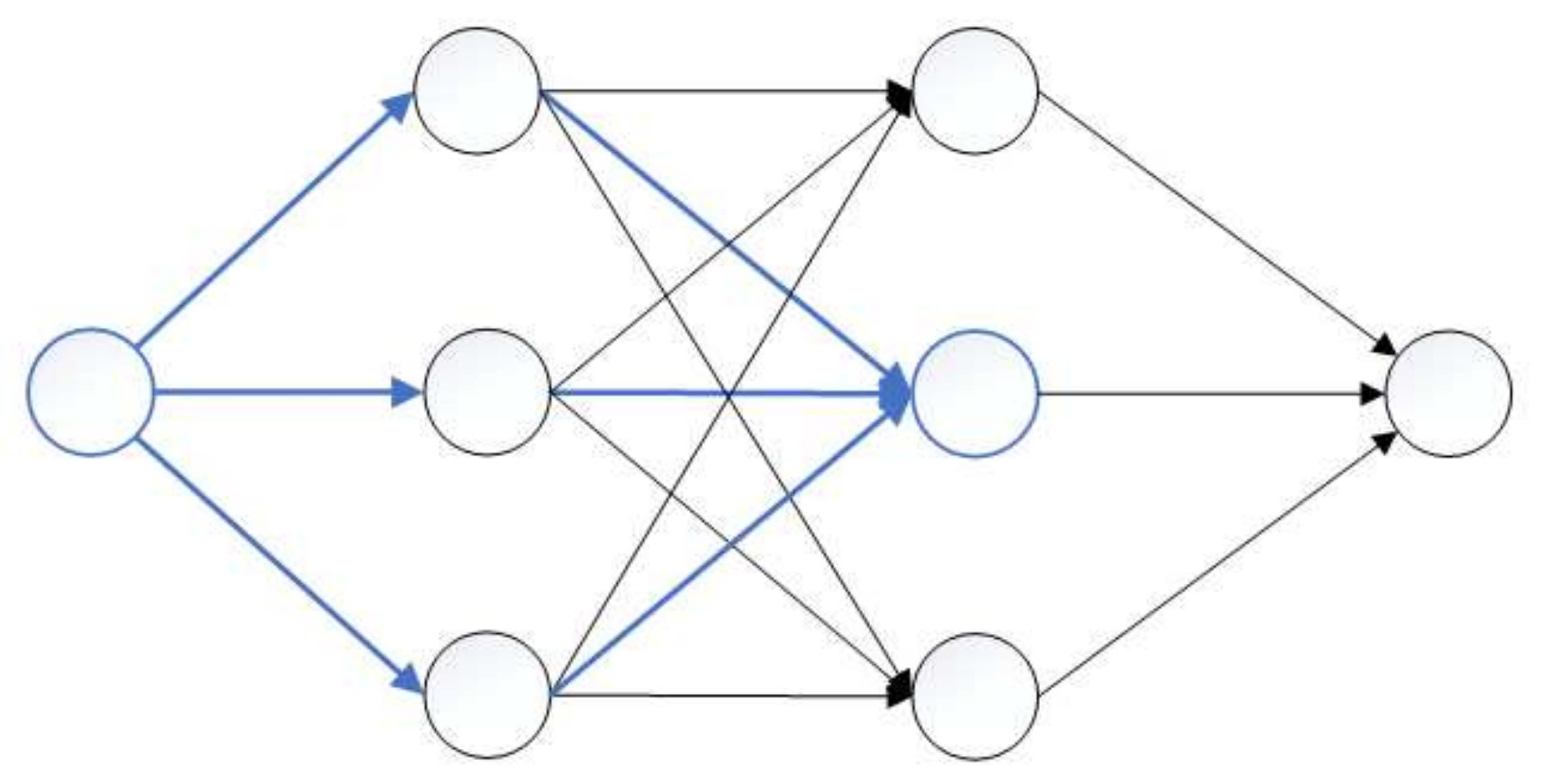}
        \caption{Interfering with oneself causes CNN to require multiple layers of convolutional layers.}
        \label{figure8}
    \end{figure}

    If we understand from the biological neural network, the wavelength of the natural color of the physicist's world changes monotonously from red to purple, but the system of human perception of color is closed-loop, such as the combination of red light and purple light is understood as the monochromatic magenta color. However, does not physically exist with a single physical wavelength of light corresponding to the color, but the human perception system fusion understands as a single color. Therefore, the convolution neural network must be fused or transformed after the input scattering (decomposition) must be multilayered.

\item[(8)] Convolution is not the extraction of human knowledge such as image edges or colors, the convolution is the physical law and does not require human prior knowledge. From (Equation \ref{equation2}) to know the image gradient is obtained after the translation operation, so for the first feature layer of the image we can see that it reflects the edge of the image. But as the subsequent continuous pooling and convolution, feature layer image will become more and more difficult to understand and abstract compared to the original layer. However, in our model, the convolution image is a diffraction stripe or patch image after destructive interference or constructive interference, which is the superposition of the same or coherent wave. That is, the images are decomposed and classified by the same coherent attribute. The wave of the same phase has a destructive interference and the opposite phase has a constructive interference, and the interference diffraction patterns will effectively deconstruct the entire information of the incident image. And this coherent property is the symmetry of the image, according to quantum mechanics, the physical quantity described by the symmetry is the wave vector. It can also be said that CNN is to measure the wave vector of the image, because it is the $XY$ plane convolution of the image, i.e. the measurement of the momentum ${p_x},{p_y}$ in the $x,y$ direction. According to quantum mechanics, the $z$ directional angular momentum operator is defined as: ${\hat l_z} = x{\hat p_y} - y{\hat p_x}$ , so it also detects angular momentum or rotational symmetry around the $z$ direction. That's to say, a trained CNN measures the translational symmetry of the $x,y$ direction of the image or the rotational symmetry of the $z$ direction through the translational operation and interaction of the image, is classified as the interference diffraction fringes or patch image that we see. The feature map layer is effectively classified according to the symmetry, and then the whole connection layer is mapped, which can effectively recognize the image.

\end{itemize}

In summary, the above research shows that LeCun's most famous contribution, the convolutional neural network, is entirely based on prior knowledge and that the idea of not requiring human structured knowledge is completely correct. CNN is based on the scientific theory, so it is the best method in image recognition. CNN is a physical model, so it is very successful in the processing of physical data such as images, videos, sounds, condensed matter physics, etc., but the performance is worse than other models when processing the strong subjectivity of symbolic modeling. For example, no matter how much data is input, it is impossible to train a model that can read product descriptions and generate an appropriate code base \cite{17}.

\subsubsection{Significance of classification layers in CNN}\label{2.2.4}

The convolutional layer is followed by a classification layer whose meaning maps the mechanical characteristics learned in the convolution layer to human symbols (marks). The significance of the classification layer is expressed in:
\begin{itemize}
\item On the significance of the classification layer, the first is mathematics. In order to effectively apply the ability of neural network that can approximate arbitrary functions mathematically, the classification layer must be fully connected.

\item The classifying layer in neural network is also physical. In our model, the neuron is a probability wave. The output of the last layer's activation function must be guaranteed to meet the normalized requirement, generally, the softmax function with a clear physical meaning is used. The activation function is shown in Section \ref{2.5}.

\item The classification must have generalization and must obey the laws of statistical physics and must cooperate with entropy. The content of entropy is shown in Section \ref{3.2}.
\end{itemize}

\subsubsection{Development prospects of CNN}\label{2.2.5}
\begin{itemize}
\item[(1)] According to the neuron scattering theory, strictly speaking, the convolutional neural network should be extended to the complex number field so that it can use the wave function with phase information. Thus, the input image is the intensity, and it should be squared root when used as a wave function, which is equivalent to the fact part:
    $$\psi (x,y) = \sqrt {I(x,y)} $$
    However, in the actual calculation of CNN, the calculation on the complex number field may not be significant. In Equation \ref{equation10}, when we compute $S$, we need to square $\psi $ , and ${\left| \psi  \right|^2}$ is proportional to the intensity of the image. The calculation of S in the real field will appear negative, but subsequent Relu rectification guarantees that the output is positive. It still conforms to the model where the value of the neuron is the intensity value, the significance of max pooling is to find the brightest neurons. In short, if the convolution is calculated in the complex number field, the square root is computed first, and then the square is computed, which is equivalent to repeating computation. And if the convolution is calculated in the real field, input do not square root, output also do not square, the Relu rectification function guarantees that no negative number will appear. By reducing the repetition calculation and saving the calculation amount, the calculation error can be reduced. There is no big difference between the calculated effect on the real field and the strict complex number field.

\item[(2)] Choosing the appropriate convolution kernel is crucial for obtaining the most significant and important information contained in the input signal. This allows the model to make better inferences about the content of the signal. The goal of this transformation is to change the data in a way that is more easily separated by the classifier. According to Equation \ref{equation16}, a better model is designed using the convolutional features, such as better coordinates: cylinder coordinates or spherical coordinates of CNN \cite{15}.

\item[(3)] The training of CNN based on quantum scattering theory obtains a large amount of input information, but the final classification only occupies a small part of the information. These are correct and objective reflections of the input information, it is easy to drown in the training process. This is undoubtedly a waste of valuable prior probabilities that can be used to migrate large-scale network knowledge into small-scale networks. Therefore, making full use of this information or data migration learning in other fields is also a topic worthy of study.

 \item[(4)] Multilayer CNN can not only perceive the translational symmetry and rotational symmetry of input, but also transform and perceive the fusion of various combinations of colors. In addition, it should also can transform and perceive the polarization direction of light. That is, CNN can perceive very rich three-dimensional scenes and behavioral information from static two-dimensional images. The geometry school will laugh : how can an image calculate three-dimensional, which is mathematically impossible. It doesn't make any sense mathematically, but it makes sense physically! Human beings can perceive three-dimensions from two-dimensional images, and deep convolution neural networks can extract 2.5-dimensional information from two-dimensional images. This is the quantum effect: White light can be decomposed into red, orange, yellow, green, cyan, blue and violet through neuronal interaction or coupling or entanglement, and random monochromatic light can be decomposed into various polarized light. According to the physical model above, a lot of new information can be found from the polarization information, such as stereo information, like the principle of stereoscopic film. This is an important prediction of the deep convolution neural network in this dissertation.

 \item[(5)]The CNN discussed above is to classify the spatial symmetry of the image by spatially translating and interacting with neurons. After the above measurement, the state after the collapse is measured as the new environment of Hamiltonian to state a new round of evolution. If the incident wave changes, such as the time dependent dispersion of the packet, the visual retention, or the Hamiltonian contains time, then it will be a $Schr\ddot{o}dinger$ equation that contained time. At this time Equation \ref{equation17} is time-containing, the green function is divided into the delayed Green function and the advanced Green function, which can be computed to predict the short future by or to recall a brief past. For time translation operations, according to quantum mechanics, the conserved quantity of time translation invariance is energy. Through the convolution of time translation, it is also possible to classify the energy properties of the input information. And the scattering of neurons contains time, so it will be an important research direction and may explain the important functions of visual retention, visual prediction and memory.

 \item[(6)]The convolution neural network based on quantum scattering theory is expected to compute the internal properties of the biological neurons and to gain more understanding of the functions of the biological neurons and the human brain.

\item[(7)] CNN is supervised training, according to the above theory, it can be used in unsupervised pre-training and training.

\item[(8)]In our neural network model, all the neuron cells of CNN form an image of a grid cell and a location cell, which is very similar to the image of the biological grid cell and position cell. It will be an important research direction to combine CNN with LSTM to achieve navigation like the brain grid cells.
\end{itemize}
	
\subsection{Deep learning based on energy model}\label{2.3}
Energy is one of the most important concepts in physics and even in the entire natural sciences. All forms of physical movement have the concept of energy, such as mechanical, electromagnetic, thermal, light, chemical movement, and biology, etc. The analysis of energy can greatly simplify the analysis of material motion. Energy is the measure of movement transformation, and it is an additive amount, following the law of conservation of energy. Energy in quantum mechanics is represented by the Hamiltonian operator.

In our model, the strength of neurons is described by the square of the wave function. The typical model of the neural network composed of interacting neurons in deep learning is the Boltzmann machine.

According to quantum mechanics, the wave function of the system is $\Psi$, and the statistical average of the mechanical quantity $O$ is:
\begin{equation}
\frac{{\left\langle {\Psi }
 \mathrel{\left | {\vphantom {\Psi  {O\left| \Psi  \right.}}}
 \right. \kern-\nulldelimiterspace}
 {{O\left| \Psi  \right.}} \right\rangle }}{{\left\langle {\Psi }
 \mathrel{\left | {\vphantom {\Psi  \Psi }}
 \right. \kern-\nulldelimiterspace}
 {\Psi } \right\rangle }} = \frac{{\sum\nolimits_x {{{\left| {\Psi (x)} \right|}^{\rm{2}}}O(x)} }}{{\sum\nolimits_x {{{\left| {\Psi (x)} \right|}^{\rm{2}}}} }}\label{equation18}
\end{equation}

There is an important conclusion between the quantum mechanics measurement and the statistical sampling of wave functions: if $\Psi$ is the eigenstate of the system energy, then:
\begin{equation}
\left\langle {\left. {{H^2}} \right\rangle } \right. = \left\langle {{{\left. H \right\rangle }^2}} \right. = E_0^2\label{equation19}
\end{equation}

That is, the variance of the Hamiltonian energy $H$ is 0, which means that if the system is in the ground state ${E_0}$ , the statistical fluctuation completely disappears. This feature is very important because it means that the closer we are to the ground state, the less fluctuations we have in the amount we want to minimize, and the less energy we have. According to quantum statistical physics, if the system is a closed system, there is energy fluctuations, but the total number of particles is constant, the multi-body probability density $\rho (x)$ satisfies the Boltzmann distribution:

\begin{equation}
\rho ({X^N}) = \frac{1}{{{c^N}{z_N}(T,V)}}{e^{ - \beta H({X^N})}}\label{equation20}
\end{equation}

Partition function is: $Z = \frac{1}{{{c^N}}}\int {d{X^N}} {e^{ - \beta H({X^N})}}$ , probability density is the normalized ${\left| \Psi  \right|^2}$ . For a neural network composed of interacting multi-body quantum systems, the general form of Hamiltonian is:
\begin{equation}
H =  - \frac{{{\hbar ^2}}}{{2m}}\sum\limits_i^N {\nabla _{{{\vec r}_i}}^2}  + \sum\limits_i {{V_1}(} {\vec r_i}) + \sum\limits_{i < j} {{V_2}(} {\vec r_i},{\vec r_j})\label{equation21}
\end{equation}

The potential energy between two particles in physics is chosen as a function of the distance between particles in the form of:
\begin{equation}
{V_{12}} = V({\vec r_1},{\vec r_2}) = V(\left| {{{\vec r}_1} - {{\vec r}_2}} \right|) = V({r_{12}})\label{equation22}
\end{equation}

Without regard to kinetic energy, only the interaction is considered, the interactions always interact in pairs, that is, the interaction energy (potential energy) takes only quadratic term, so the energy model is obtained:
\begin{equation}
E(v,h) =  - {b^T}v - {c^T}h - {v^T}Wh\label{equation23}
\end{equation}

Where $b$,$c$ and $ W$ are all unconstrained, real-valued learning parameters. The model is divided into visible layer $v$ and hidden layer $h$, and the interaction between them is described by matrix $W$ \cite{18}.

This is the most basic component of deep learning¡ªthe Restricted Boltzmann Machine energy model. It is consistent with the view put forward in this dissertation that neural network in deep learning is a physical system and conforms to the laws of physics. For Boltzmann distribution sampling, the method of random gradient descent that minimize the energy can determine network parameters  $b$, $c$ and $ W$.

\subsection{Relationship between energy model and convolution model}\label{2.4}
From the above analysis, CNN is based on accurate quantum physics, and RBM is based on statistical physics which will be described later, the difference is pure ensemble and hybrid ensemble.

The traditional RBM can only express the probability distribution function with a positive value. In order to make it suitable for describing the wave function with phase information, Carleo and others extend the parameters of RBM to the complex field. In addition, in the actual calculation, the function form used by Carleo is the product of multiple RBM that share weights. This structure is equivalent to a single hidden-layer convolutional neural network, thus ensuring the spatial translation invariance of the physical system in the structure \cite{19}.

In addition, the RBM ignores the effect of the kinetic energy of the neuron, that is, the centroid motion, or ignores the quantum effect of the neuron itself, so the effect of the model is obviously worse than that of the CNN.

\subsection{Normalization of neural networks}\label{2.5}
It has been explained that the wave function of the neural network system is $\psi$ . According to quantum mechanics, the meaning of the neuron is that each neuron represents the probability ${\left| {{c_n}} \right|^{\rm{2}}}$ that eigenvalue ${F_n}$ can be measured by a group of mechanical quantities $F$ , and the wave function satisfies the normalized condition, so $\sum\limits_n {c_n^2}  = 1$.

Once the wave function of the system is determined, the average value of the mechanical quantity and the probability distribution of the measurement result can be obtained through the mechanical quantity measurement. The results obtained by the physical system are obtained by measurement, whereas in quantum mechanics, measurements differ from those in classical mechanics. The quantum measurements affect the measured subsystems, such as changing the state of the measured subsystem, which called the wave function collapse, and the measured result accords with a certain probability distribution. In other words, when the mechanical quantity is measured, the state is collapsed to the eigenstate of this mechanical quantity, then the mechanical quantity is its eigenvalue. But it may also be other eigenvalues, that is, it may also be collapsed to other momentum eigenstate, which is a probability, so it must have:
\[\sum {_i} h_i^l = 1\]

That is to meet the normalization conditions. Quantum measurement is the core issue of the quantum mechanics interpretation system.

In deep learning, considering the concept of mechanical quantity measurement, the result output of each layer neural network characteristic (eigenvalue) learning should be a quantum measurement, so the activation function is needed to determine its probability distribution. If there is no activation function, then the network can only express the linear mapping. Even if there are more hidden layers, the entire network and the single-layer neural network are equivalent. The meaning of the activation function is the output of the quantum measurement, which determines its probability distribution and is normalized. The most important activation functions are:
\begin{itemize}
\item $\displaystyle softmax({h_i}) = \frac{{\exp ({h_i})}}{{\sum\nolimits_j {\exp ({h_j})} }}$.

\item $\displaystyle sigmoid({h_i}) = \frac{1}{{1 + \exp ( - {h_i})}}$

\item $Relu({h_i}) = \left\{ \begin{array}{l}
 {h_i},({h_i} > 0) \\
 0,({h_i} \le 0) \\
 \end{array} \right.$
\end{itemize}

The first activation function guarantees that the sum of all output neurons is 1.0, which guarantees the normalization of its probability distribution. So the last layer basically apply this activation function. The softmax function has a clear physical meaning, which is the Boltzmann probability distribution of the ideal gas or quasi particle.

The sigmoid activation function is more commonly used in traditional neural networks, but it is not suitable for deep neural networks. A major breakthrough in deep learning technology is the use of a third activation function. The Relu function directly outputs a probability or probability density distribution which is a quasi-particle number in our physical image, and it must be greater than or equal to 0.

\section{A macroscopic view of deep learning }\label{3}
The deep neural network is an interacting quantum multibody system, according to quantum mechanics, by determining the wave function p of the quantum multi-body system, the whole information of the quantum system is grasped. The core goal is still to find the solution $\psi $ of the $Schr\ddot{o}dinger$ equation, this is generally very difficult. However, because we are mainly concerned with the macroscopic properties such as statistical mean of system observations, macroscopic performance is subject to statistical physics. Statistical physics is a bridge between microscopic and macroscopic, which determines the condition of macroscopic equilibrium state and stable state, and the change direction of macroscopic state. Quantum statistical physics is described in two different ways, pure ensemble and hybrid ensemble, so there also are two different kinds of neural networks.

Ensemble is an abstract concept, which is divided into pure ensemble and mixed ensemble. If the $N$ subsystems in the ensemble are all in the same state  $\left| \psi  \right\rangle $, then the ensemble is called pure ensemble and they are described in pure state $\left| \psi  \right\rangle $ . If $N$ systems have ${N_1}$ systems in state $\left| {{\psi _{\rm{1}}}} \right\rangle $, ${N_2}$ systems are in state $\left| {{\psi _2}} \right\rangle $ , ...   ${N_i}$ are in state $\left| {{\psi _i}} \right\rangle $ ,..., the probability of each measurement system being in $\left| {{\psi _{\rm{1}}}} \right\rangle ,\left| {{\psi _2}} \right\rangle , \cdots ,\left| {{\psi _i}} \right\rangle , \cdots $ state is ${P_1} = \frac{{{N_1}}}{N},{P_2} = \frac{{{N_2}}}{N}, \cdots ,{P_i} = \frac{{{N_i}}}{N}, \cdots ,$ and the set of $N$ systems($N\to\infty $)is called a hybrid ensemble, which is described by a set of all $\left| {{\psi _i}} \right\rangle $ and ${P_i}(i = 1,2, \cdots )$ .

\subsection{Purely ensemble and hybrid ensemble neural networks}\label{3.1}

The theory of quantum statistical physics is used to design the neural network architecture, which has two kinds of neural networks, pure ensemble and mixed ensemble. The practical applications of deep learning generally have both pure ensemble and mixed ensemble neural network structures.

Pure ensemble is described by the pure state wave function, so the pure ensemble neural network architecture goal is to find the connection weights of the neural network through training so that the pure state can be constructed together with the input, and each layer can obtain the probability of each value (characteristic quantity) of the observed amount. What it pursues is a numerical solution of the wave function.

The neural network structure of hybrid ensemble is suitable for using unsupervised training to find the set---mixed state, which does not correspond to an observation. and is designed to reconstruct input. It is not from the Schr?dinger equation of quantum mechanics to find the wave function of the system. Instead, it believes that under certain macroscopic conditions, at certain moments the system will be in a quantum state with a certain probability. The hybrid ensemble requires secondary statistics, which can¡¯t be solved in quantum mechanics, so wo need to introduce additional basic assumptions of statistical physics---when the isolated system reaches equilibrium, the probability of appearance of various microscopic states is equal. The macroscopic quantity of the system is the statistical average of the various quantum states that the system may be in according to the corresponding microscopic quantity.

One important technical difference between these two neural network structures is:

Pure Ensemble only needs to find out the probability distribution ${\left| {{c_n}} \right|^{\rm{2}}}$ of each layer's observable amount through supervised training, it can satisfy the requirement of human application, do not have to perform layer-by-layer pre-training, and each individual layer is entirely determined by quantum mechanics. From Equation \ref{equation13} we know that the process is a physical calculation and the result is a possible microscopic state, whether this microscopic state is a macroscopic state that satisfies the requirements will be determined by supervised training at all levels. The deep learning of hybrid ensemble, because there is no computational formula of quantum mechanics like Equation \ref{equation13}, it must be pre-trained layer by layer to find the probability distribution ${\left| {{\psi _i}(x)} \right|^2}$ of the various states of the ensemble, then all layers should have supervised training to find the probability distribution P of the system¡¯s state ${\psi _i}(x)$ , so there must be a second statistical calculation. The size of P can¡¯t be solved by quantum mechanics, it requires statistical mechanics principles or assumptions. This answers Yoshua Bengio's perplexity in the paper \cite{20}: training depth-supervised neural networks is usually very difficult without pre-training with unsupervised learning, but the exception is CNNs. So we are very curious about what special point in the structure of CNN make it has very good generalization performance in tasks such as image processing? The answer is that CNN is the purely ensemble of neural network structures, as explained earlier, its individual layers are completely described by quantum mechanics. The hybrid ensemble of neural networks that must be used for unsupervised learning layer-by-layer pre-training, such as the RBM described previously.
	
\subsubsection{Pure ensemble neural network}\label{3.1.1}

The pure ensemble neural network are similar to instrumentation, belongs to the category of physical measurement, so its learning performance can be checked because of its clear physical meaning.

The pure ensemble neural network has a prominent advantage. The model it trains is likely to be a general model, sometimes it is not only useful in this field, but also can be used across fields. For example, the CNN trained on ImageNet can also be used for image recognition in furniture projects, as well as for a wide range of computer vision applications, and can effectively serve as a general model for our visual world. However, as the network deepens, the versatility of the neural network away from the input layer will be significantly reduced, and it is generally not reusable. The reusability depends on the depth of the model, because measurement is essentially the interaction of the input signal and the neural network. As the layer increases, the more damage to the system, the stronger the specificity of the measurement results and the less versatility. For example, in a specific CNN, layers appearing earlier in the model will extract locally highly generalized feature maps, while higher levels will abstract more abstract concepts (such as ¡°cat ears¡± or ¡°dog eyes¡±). Therefore, if your new dataset differs significantly from the dataset trained by the original model, it is better to use only the first few layers of the model for feature extraction instead of using the entire convolution model.

The pure ensemble neural networks are completely determined by quantum mechanics, so it is bound to have a good effect once applied. For example, using CNN to recognize the image is the best, and can be widely applied to the front end of various types of neural networks, and can be widely used in the front-end of various types of neural networks.

\subsubsection{Hybrid ensemble neural network}\label{3.1.2}

The hybrid ensemble neural networks are neural networks such as classifier, generator, automatic encoder decoder, and the like. The hybrid ensemble neural networks and pure ensemble neural networks are often opposite in their characteristics, and the well-trained hybrid ensemble neural networks are generally not reusable; the performance of learning is not checkable, and it is essentially a black box, because the view of the wave function does not have any macro meaning. For example, the information that the classifier in image recognition learns about the probability of existence of a class in the entire graph, completely frees from the concept of space. For issues where the location of the object is important, densely connected characteristic will be essentially useless.

In hybrid ensemble, the observations of mechanical quantities are the results of two statistics. The first is quantum mechanics, which is caused by the probability property of the state vector; and the second is due to the mixture of states, which is caused by the statistical nature of the state at which the system may be. So the hybrid ensemble of neural networks must be pre-trained with unsupervised learning to meet the probability distribution of the state vector of quantum mechanics, and then use supervised learning to fine tune the entire network (second statistic) to determine macroeconomic equilibrium or stable state.

\subsection{Understanding deep learning from the concept of entropy}\label{3.2}
\subsubsection{The role of entropy in deep learning}\label{3.2.1}

Information is not a simple mathematical concept, but a basic physical concept like matter and energy. Therefore, all the processing of information (such as computing) is subject to the basic laws of physics, and a key concept in information theory is entropy. Thermodynamic entropy, which is well known to physicists, is homologous to the information entropy that Shannon uses to measure information. The physical metric that describes the number of system states is the entropy. For example, Brownian motion is the irregular motion exhibited by tiny particles, and the higher the temperature, the more violent the irregular motion of the molecules. From the microscopic point of view, Brownian motion is disorganized, but from a macroscopic point of view, it is a macroscopic law with irreversible (increased entropy). Therefore, the entropy points out the direction of the evolution of the system and describes the conditions under which the system is in equilibrium. In deep learning, a large amount of data is needed to train the discovery system model. Different models have different entropy, the number of all microscopic states corresponding to the correct model is extremely large (namely, the entropy is maximum). The cost function of correlated entropy in deep learning applications is defined based on the entropy of the physical principle.

The definition of entropy in physics is: $H = k\ln \Omega $ , where $\Omega $ is the number of microscopic state. Entropy is additive, the entropy of system is the sum of the entropies of each subsystem, and entropy is a homogeneous function. Let $X$ be a discrete random variable whose probability distribution is $P(X = {x_k}) = {p_k},k = 1,2,3, \cdots ,n$ , the probability of occurrence of ${x_k}$ is ${p_k}$ , then the number of microscopic state is $1/{p_k}$ , so the cumulative sum of all the entropy of the random variable $X$ is:
\begin{equation}
H = \sum {{p_i}\ln 1/{p_i}}  = \sum {{p_i}\ln {p_i}} \label{equation24}
\end{equation}

The probability distribution is $P(x)$ , and the random variable is the number of microscopic states of $X$, i.e. the entropy is:
\begin{equation}
H(X) =  - \sum\limits_{k = 1}^N {P(X = k)\log P(X = k)} \label{equation25}
\end{equation}

Cross entropy is an extended concept of entropy, it introduces a second probability distribution. Then the number of microstates (entropy) for a random variable $X$ is:
\begin{equation}
H(p(x),q(x)) =  - \sum\limits_{x \in X} {p(x)\log } q(x)\label{equation26}
\end{equation}

Cross entropy measures the number of microstates of a random variable X under these two probability distributions using physical quantity entropy. Cross entropy contains the difference between two probability distributions, and when the two distributions are the same, the entropy is Equation \ref{equation25}. It is used in deep learning applications to measure the degree to which the model distribution approximates the unknown distribution.

Therefore, entropy plays an important role in deep learning, it has become the objective function of the selection and adjustment of the parameters of the deep learning model:

\begin{itemize}
\item How the neural network adjusts parameters, in which direction the parameters are adjusted, the basis is the entropy of the random variables.

\item How to determine the conditions for the physical balance and stability of the neural network, the basis is the entropy of the system. For example, the entropy of an isolated equilibrium system must be maximum.

\item Entropy allows us to precisely define " correlation " and extract features from massive data.

\end{itemize}

\subsubsection{Application of entropy in cost function}\label{3.2.2}
According to statistical physics, a macroscopic state with a steady state and balance is a state where the number of microscopic states is maximum (i.e. the entropy is maximum). This is the physical basis for the selection of the cost function.

The deep learning neural network discussed above are divided into two parts. The former is the neural networks that conforms to the laws of physics, such as each convolution layers of CNN, not only the characteristics of the trained convolution layer have physical significance, each of the conditions in the training also has physical significance. Because it is calculated according to the physical equations, except that this microscopic state does not correspond to the final trained macroscopic state, or the number of microscopic states that the final trained macroscopic state has is very small (entropy is small). The latter part is the classification of neural networks, such as a convolution layer that then maps human symbols (labels) to their physical characteristics. This is a distribution of two completely different concepts, so the cross-entropy of the model's output (physical) and the training target (labeled) should be used as a cost function to approach the real physical model. The magic of the wonderful combination of classification problems and cross-entropy is that even if the cross-entropy on the test set is over-fitted, the classification error will not be overfit \cite{21}.

For the regression problem, it is the same kind of problem, so the mean square error is often used as the objective function or the cost function, which is equivalent to the general entropy maximum. KL divergence is closely related to cross entropy, but it does not include the entropy of the model, it reflects the difference between the two distributions. Unlike the use of cross entropy, the KL divergence is used when there is no more physical distribution. These analyses show the powerful power of statistical physics in understanding deep learning.

\subsection{Understanding deep learning from the concept of master equations}\label{3.3}

The evolution equation of the Markov process probability distribution is the master equation, which is one of the most important equations in statistical physics because it is almost universally applicable. The Markov process is a process in which most of the memory effects can be omitted during the evolution process. When the system evolution of a random variable occurs, the transfer process will occur between different values of the random variables, by transferring the probability of the system changes in a given state until the system reaches a final equilibrium state.

In deep learning, one of the simplest examples of the Markov process is the Markov chain. The Markov chain is defined by a random state ${X_t}$ and a transition distribution $T({X_{t + 1}}|{X_t})$ . The transition distribution $T$ is a probability distribution showing the probability of a random transfer to ${X_{t + 1}}$ in the case of a given state ${X_t}$ . Running a Markov chain is the process of constantly updating the state ${X_t}$ based on the value ${X_{t + 1}}$ of the transition distribution $T$. Among them, the probability distribution of the system state at time $t+1$ is only related to the state at time $t$, and has nothing to do with the state before $t$.

\subsection{Understanding deep learning from the concept of renormalization group}\label{3.4}

David Schwab and Pankaj Mehta found that the Deep Belief Network (DBN) invented by Hinton resembled the renormalization in physics in a specific case, which means that the details of the physical system are obtained in a coarse-graining manner to calculate its overall state. When Schwab and Mehta applied DBN to a magnetic model at the ¡°critical point¡± (when the system is fractal and self-similar at any scale), they found that the network automatically uses a reorganization-like process to discover the state of the model. This discovery is shocking, as the biophysicist Ilya Nemenman commented, it shows that ¡°extracting related features in the context of statistical physics and extracting related features in the context of deep learning are not just similar, but they are completely the same."\cite{22}.

The renormalization group is a mathematical tool for examining changes in the physical system at different length scales. The details of the physical system are physically obtained in a coarse-grained manner to calculate its overall state. The important feature of this method is that it is independent of the system type. Each key step in the renormalization group method is based on the main characteristics of the system, rather than putting the system into the framework we are familiar with, and then adjusting the parameters \cite{23}. The standard renormalization process in statistical physics is equivalent to the feature extraction process of supervised learning in depth learning. The information is transmitted layer by layer and eventually converges to the theoretical boundary (fixed points). The purpose of the renormalization group method is to obtain new features, and to ensure that the Hamiltonian function forms are unchanged under the new scale of renormalization.
\begin{figure}[H]
    \centering
    \includegraphics[width=3in]{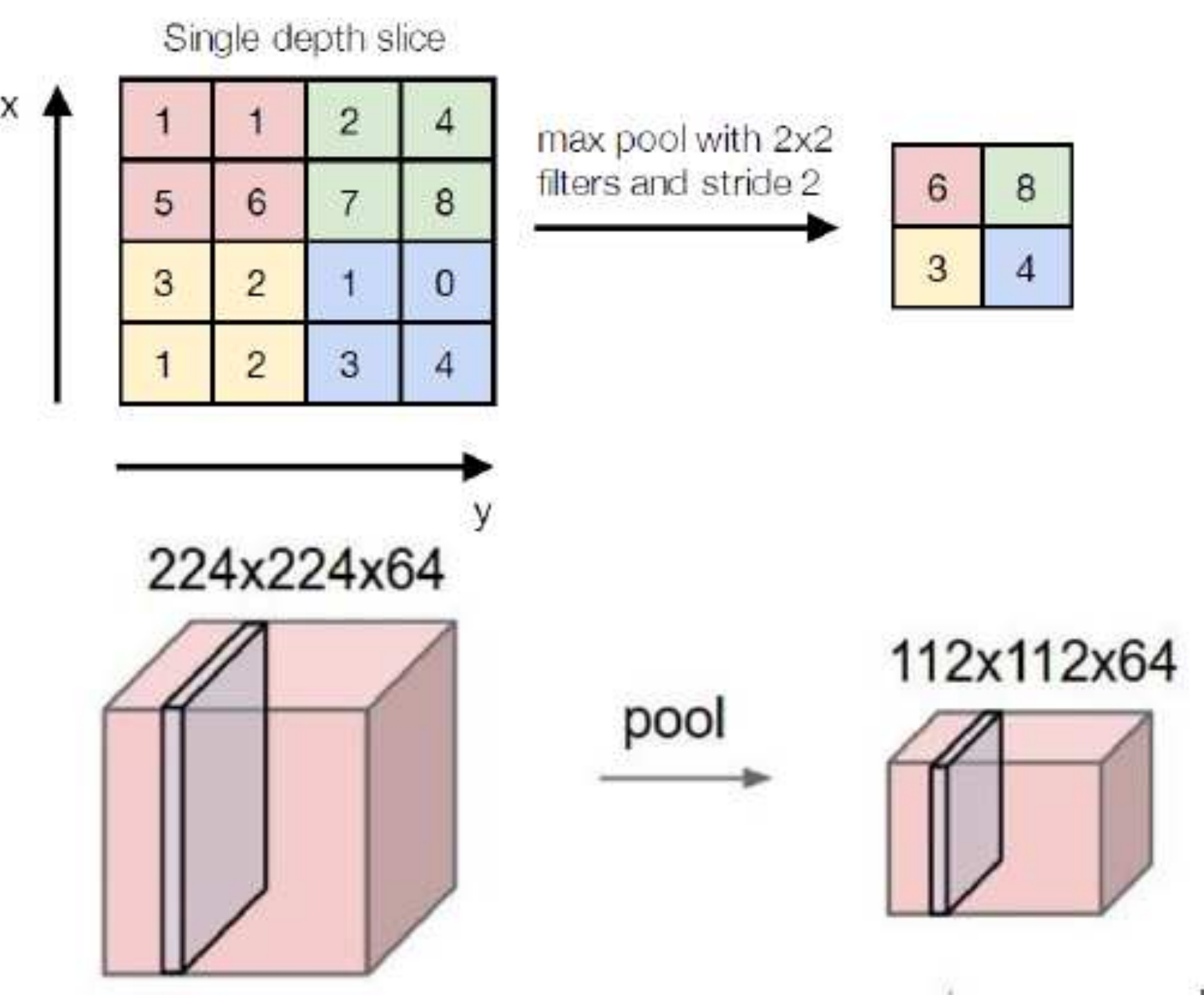}
    \caption{The pooling operation in convolution neural network.}
    \label{figure9}
\end{figure}

For example, the pooling operation (pooling layer) in CNN is based on the main characteristics of the system and uses the max pooling method to integrate the feature points in the small neighborhood according to self-similarity to obtain new features. As shown in Figure \ref{figure9}, the pooling layer uses the max pooling (where the size of the filtering core is 2*2 and the step size is 2) to fuse a feature with an input size of 4*4 and retains only the largest feature point in the area, then the characteristic size after the pooling operation is 2*2, and the pooling layer plays a role of dimensionality reduction.

It has been proved that there is a one-to-one mapping relationship between RBM-based deep neural networks and variational renormalization groups. The paper~\cite{7} illustrates the mapping relationship by analyzing the DNN and numerical two-dimensional Ising model of a one-dimensional Ising model, and it finds that these DNNs self-realize a coarse graining process, i.e. Kadanoff block renormalization. The results show that deep learning may adopt a generalized renormalization group class scheme to learn relevant features from the data. The paper proved that deep learning is closely related to the renormalization group, one of the most important and successful technologies in theoretical physics.

\section{A physical world view of deep learning}\label{4}
\subsection{The interpretability of deep learning}\label{4.1}
The interpretability of deep learning models can be divided into the following categories:

\begin{itemize}
\item[(1)] \textbf{Feature attribution} VS \textbf{Internal logic:}

The former maps the behavior of the model back to the original set of input features (or artificially creates optional input features). In the complex decision-making process of the model, the larger the influence of the characteristics will be assigned to the larger weight, the structure of human knowledge plays a decisive role in this model; The latter argues that: In the process of obtaining the final answer of the model, it is the abstract role of the physical meaning of the model itself and the internal working logic, rather than human structural knowledge. Obviously, the interpretation of deep learning in this dissertation belongs to the latter. This paper analyzes deep learning from the internal logic according to the principle of physics, while most of the papers use the former method to explain deep learning.

\item[(2)] \textbf{Simulation acquires knowledge} VS \textbf{ Introspection acquires knowledge:}

Knowledge based on simulation means that we obtain an understanding of our own model by generating some form of simulation data, capture how the model represents these data points for understanding; Introspection acquires knowledge comes from the fixed orientation of the model and use them to gain knowledge without having to simulate the former. Obviously, the interpretation of deep learning in this dissertation belongs to the latter, while most papers belong to the former, the focus of their interpretation is to visualize the characteristics of deep learning. However, this dissertation holds that the deep neural network data has high dimensional data characteristics, and human beings cannot understand the visual characteristics of high-dimensional data.

\end{itemize}

According to quantum statistical physics, a system with $s$ classical degrees of freedom, the dynamic state of the system is determined by the wave function:
\[{\psi _{{\sigma _{\rm{1}}}{\sigma _{\rm{1}}}, \cdots }}({q_1},{q_2}, \cdots {q_s},t)\]

Where $q$ is the classical coordinate and $\sigma $ is the non-classical degree of freedom. People think that deep learning is incomprehensible. It is precisely because it uses the $\psi $ to characterize the macroscopic nature, which cannot be understood through visualization, nor can it be understood through mathematics. Deep learning is the interpretation of observed data using the dynamic characteristics of microscopic layers that are not observed by humans or that are not intuitively understood.

\subsection{Locality}\label{4.2}
One of the deepest principles of physics is locality, that is, things directly affect their surroundings. Locality is a relative concept, usually referring to the scope and degree of influence of a physical quantity. Locality has two effects:

\begin{itemize}
\item[(1)] Short-range effect:

The interaction is only a nearby function, ignoring the effect of the remote, that is, the effect of the remote is averaged or canceled. For example, it has been explained before that during the operation of the Markov chain, the system state at the current moment is only related to the state at the previous moment, has nothing to do with the state before the previous moment, that is, the short-range effect.

\item[(2)] Local coupling:

The local roles of each regions are coupled and then coupled as localization. Scale changes, renormalization, long procedures, strong correlation, and coarse-graining. For example, pooling in CNN has already been described in the previous section. By using max pooling, similar features of neighboring regions are merged (that is, processes that are locally coupled), thereby achieving the effect of dimensionality reduction.
\end{itemize}

\subsection{Symmetry}\label{4.3}
Symmetry and conservation law are the basic laws of nature. Symmetry can not only reduce the number of parameters, but also reduce the computational complexity. See the convolutional neural network under the symmetry (translation invariance) we introduced in Section \ref{2.2}. Whenever Hamiltonian obeys a certain symmetry, that is, invariable under some transformation, the number of independent parameters required to describe it decreases further. The most important mechanical quantities of atoms are momentum and angular momentum, so the corresponding translation and rotation transformations are the most important and basic operations. For example, many probability distributions are constant in the case of translation and rotation.

If the system has translational symmetry, then the state after the translation operation differs from the original state by a maximum of one phase factor, that is, the difference between the input state and the output state mainly appears as a phase shift. The input wave function can be labeled by the wave vector, and the physical meaning of the wave vector is momentum, which reflects the spatial symmetry of the system. The state with translational symmetry is the eigenstate of the momentum operator, and the momentum represented by the wave vector is the corresponding eigenvalue. Therefore, the process of learning the eigenvalue and the eigenvalue is the process of obtaining the wave vector.

\subsection{Conjugacy and Duality}\label{4.4}
Conjugacy is the amount of pairing. For example, in physics: pressure and volume, temperature and entropy, intensity and extension are conjugate quantities, in mathematics: real and imaginary numbers, transposed conjugates of matrix, and so on. Duality is the correspondence between the different physical theories that lead to the same physical results, such as, there is a reaction force with acting force, there are holes with electrons. These are the important organizations of physics, and they are related to each other and to each other in physical relations. This worldview is also used in deep learning. Here are two examples:

(1) \textbf{Stochastic gradient descent algorithm using momentum}

Although random gradient descent is a very popular optimization method in the training process, the learning process is sometimes very slow, and even cannot found the best point. The basic physical idea to solve this problem is to introduce conjugates and use duality to solve this problem.

The cost function has a gradient $g$, there is flow (or velocity or momentum), so the introduction of the momentum variable $v$. The gradient of the cost function is seen as a force that pushes the cost function to accelerate downhill and the momentum increases; if there is only this kind of force, the optimization process can never stop, so we need to add another force or gradient (called the viscous resistance) to make the cost function converge to a minimum; but the minimum may not be the smallest, so wo need to increase a momentum, let him out of the local minimum, so as to achieve the overall optimal. Update the algorithm as follows:
\[\begin{array}{l}
 v \leftarrow \alpha \nu  - \varepsilon g \\
 \theta  \leftarrow \theta  + \nu  \\
 \end{array}\]

\begin{figure}[H]
    \centering
    \includegraphics[width=3in,height=2in]{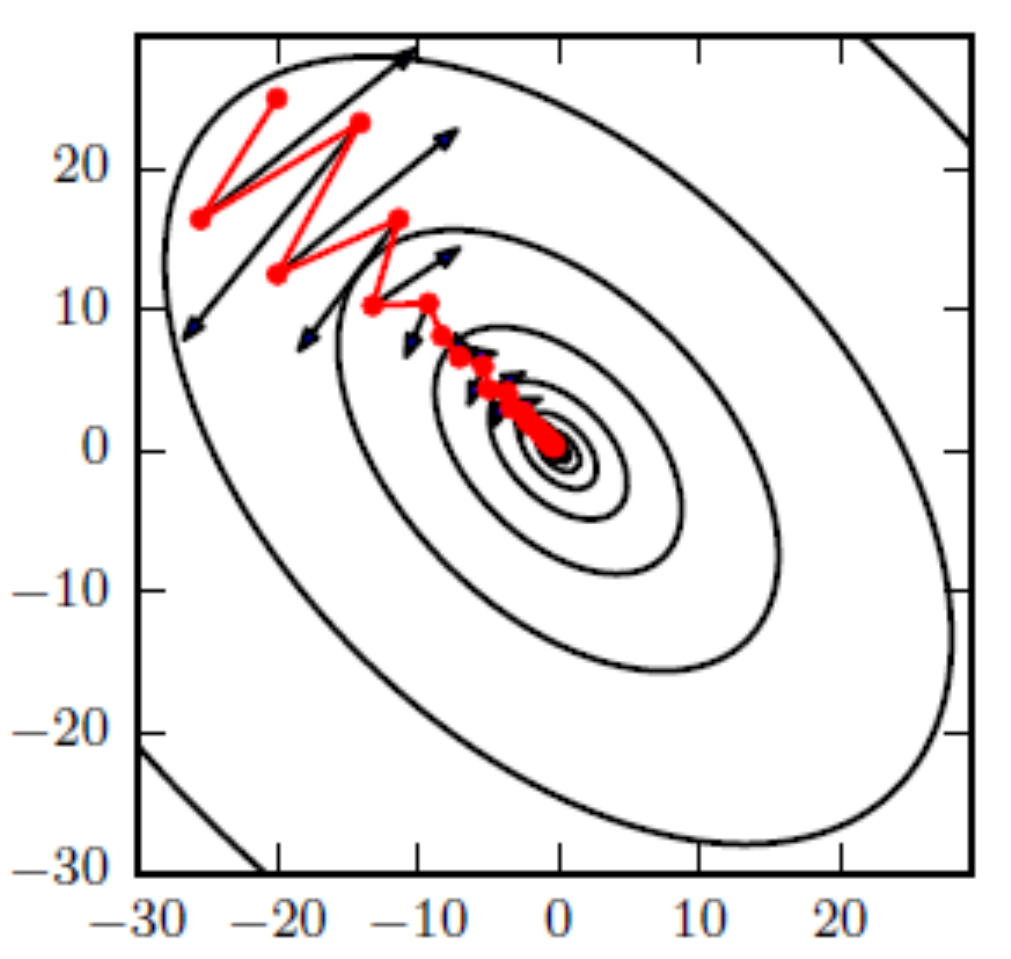}
    \caption{Stochastic gradient descent algorithm using momentum. Figure courtesy: \cite{12}}
    \label{figure10}
\end{figure}

 As shown in Figure \ref{figure10}, the contour depicts a quadratic loss function (Hessian matrix with ill-conditioned conditions). The red path across the profile represents the path followed by the momentum learning rule, which minimizes the function. We draw an arrow at each step of the path, indicating the step the gradient will take at that point. We can see that the quadratic objective function of a pathological condition looks like a long and narrow valley or a canyon with a steep edge. The momentum passes right through the gorge, while ordinary gradient steps waste time moving around the narrow axis of the canyon \cite{12}.

(2) \textbf{Annealing and tempering}

Temperature and entropy are conjugate quantities, when the temperature equals 0, the entropy equals 0(minimum); the temperature infinity is the most chaotic, the model becomes uniform distribution. Using the temperature in the Boltzmann distribution, tempering: The Markov chain temporarily samples from a high temperature distribution and returns to sampling at unit temperature. Annealing: The Markov chain temporarily samples from a low temperature distribution and returns to sampling at unit temperature to find the best among the different peaks. However, it is necessary to pay attention to the existence of a critical temperature.

Most of the models used for prediction in deep learning generally use the softmax activation function to assign probability distributions to tags. The reference \cite{24} mentions that, in the image classification problem, we divide the pictures into cats, dogs, and tigers three kinds. In a training, the probability of the three types is [0.0010, 0.0001, 0.9989], and the one-hot code of [0, 0, 1] is obtained as the classification result (hard-target). However, the intrinsic link between cats and tigers can easily be overwhelmed during training. This is undoubtedly a waste of valuable a priori probabilities that can be used to transfer large-scale network knowledge into small-scale networks. In order to make full use of the correlation between such class categories, we need to change the probability distribution in some way to make it more smooth. In the reference, Hinton further modified the softmax function:
\[{q_i} = \frac{{\exp ({z_i}/T)}}{{\sum\nolimits_j {\exp ({z_j}/T)} }}\]

When the temperature $T=1$ in the equation, it degenerates into the traditional softmax function; when $T$ is infinite, the result approaches $1/C$, that is, the probabilities on all classes approach to equal; when $T>1$, we can obtain the soft target label. By increasing the temperature T, the mapping curve of the softmax layer will be smoother, so the probability mapping of the instances will be more concentrate and the goal will be "soft". Therefore, in order to make full use of the correlation between such class categories, the method of changing the probability distribution is to increase $T$.

\subsection{Hierarchy}\label{4.5}
One of the most striking features of the physical world is its hierarchy. Spatially, it is an object hierarchy: elementary particles form atoms, then form molecules, cells, organisms, planets, the solar system, galaxies, and so on. Complex structures are usually layered and created through a series of simple steps. The observables of world are inherently hierarchical and cannot be commutative.

The main reason for the success of deep learning is the hierarchical nature of neural networks (deep neural networks). In order to extract uncomplicated features, a multi-layer neural network is required to stack simple networks and then effectively implement the generation process through layering and combination. Multilayer networks provide layer-by-layer abstract channels from low to high levels.

\subsection{Model Reusability}\label{4.6}
As a physical model, especially the convolution neural network, the deep neural network can be used as a universal method because of its physics, so it is reusable in any field. The important basis for judging whether a model can be reused or reused is its physical properties. The following analyzes the problem of model reusability of convolutional neural networks.

\begin{itemize}
\item[(1)] The convolutional network used for image classification consists of two parts: they start with a series of pooling and convolution layers and end up with densely connected classifiers. The first part is called the "convolution base" of the model, it learns completely from the physical characteristics, so its model can be used in various fields, but the dense connection layer behind is the mapping between annotation and physics, and its model is not reusable. So in practice, "feature extraction" simply use the convolved basis of the previously trained network, run new data through it, and train a new classifier over the output.

\item[(2)] There are two conditions for the reusable base to be reused. First, the input and output are the probability waves in the physical sense. Second, it internally measures the wave vector of the input physical quantity. According to the uncertainty principle of quantum mechanics, the coordinate and wave vector cannot be measured at the same time. Therefore, its output does not contain the input position information, it is impossible to obtain accurate position information of the original input image. And the position is very important for the object position, the dense connection is more completely useless.

\item[(3)] Feature characterization extracted using multi-layer architecture will evolve from simple and local to abstract and global in structure, but because of the process, the further away from the input layer the convolutional base is, the lower the reusability is. On the one hand, because in the continuous pooling, the specificity of features is getting stronger, and more and more physical information is lost. On the other hand, equation \ref{equation8} is calculated using the input as the wave function, the error is getting bigger and bigger, it deviates more and more from the real physical model, so the reusability is getting lower and lower. So fine-tuning is another widely used model reuse technique. Fine-tuning freezes convolutional groups near the input layer without training, thaws convolutions away from the input layer, and train with the new classifier to make them more relevant to the problem at hand.

\end{itemize}

\subsection{Model vulnerability}\label{4.7}
Although deep learning has achieved great success in many applications, there are still many limitations: for example, it needs a lot of data, the vulnerability of the algorithm etc. Why is the neural network easily disturbed by the input of small disturbances? From the physical model of deep neural network in this paper, we can see that on the one hand this is a physical problem, behind which we have not found new physical phenomena or new physical modes; On the other hand, labeling does not match the actual physical characteristics, such as cross-entropy, the two distributions cannot be approximated, or they are approached in a group of training data, it does not mean that the test data can match nonexistent labels, especially the deception problem. Therefore, the key to the vulnerability of the model is the limitation of human understanding. According to the physical world view of solving problems, providing anti-classification is a direction worthy of study.

\subsection{Causality and Correlation}\label{4.8}
At present, deep learning pays attention to relevance instead of causality, and uses joint probability distribution to replace traditional theorems and laws. The theoretical foundation of deep learning methods lies in the representation and transformation of statistical probability distributions. It is consistent with the physical model of the deep neural network in this paper. That is, the state of the microscopic particle is completely described by the wave function $\psi $ . After the wave function is determined, all the characteristics of the system can be obtained, the average value of any mechanical quantity and its measurement possible value and the corresponding probability distribution are also completely determined.

However, in practice, a mathematical model of causality derived from statistical data will be comprehensively used. For example, these models can be used to establish a causal relationship between smoking and cancer, or to analyze the risks of a construction project, and so on. Can these mathematical models be extended to the microscopic world dominated by quantum mechanics? Can it be incorporated into the deep learning quantum physics model? Since quantum mechanics itself has many strange features, for example, if two or more quantum systems are entangled with each other, it is difficult to deduce whether the statistical correlation between them is causal.

The concept of causal information actually exceeds statistical relevance. For example, we can compare these two sentences: "The number of cars is related to the amount of air pollution" and "The car causes air pollution." The first sentence is a statistical statement, and the latter sentence is a causal statement. Causality differs from relevance because it tells us how the system will change under intervention. In statistics, causal models can be used to extract causality from the empirical data of complex systems. However, there is only one component in a system of quantum physics¡ªthe wave function $\psi $ , so mathematical models that use causality derived from statistical data cannot be applied, including Bayesian inference. John-Mark Allen of Oxford University in the United Kingdom proposed a generalized quantum causal model based on Reichenbach's principle of common reason\cite{25}, successfully combining causal intervention and Bayesian inference into a model.

\section{Conclusions}\label{5}
At present, the research on the internal theory of deep learning is very scarce, and the successful application of deep learning and the limitations of its existing technology further illustrate the importance of studying its internal technical mechanism from a scientific perspective. Only knowing why to do it can transform existing methods or means from a deeper level. This is the scientific way of thinking. Based on the principles of physics, this paper interprets the deep learning techniques from three different perspectives: microscopic, macroscopic, and physical world perspectives. Inspired by the biological neural network, a new neuron physics model was proposed. Based on this, it explains the success of deep learning well, and fully reveals the internal mechanism of deep learning by scientific methods. A good theory can not only explain existing experiments, but also predict new phenomena and technologies. Therefore, this dissertation also proposes the direction of further research in deep learning. Some of the main conclusions of this paper are as follows:

\begin{itemize}
\item[(1)] The deep neural network is a physical system, and its architecture and algorithm should conform to the principles of physics. The technical foundation for deep learning is physics, especially quantum physics and quantum statistical physics.

\item[(2)] In this dissertation, the physical meaning of neurons in deep neural network is proposed: its output value is the distribution of quasi-particles.

\item[(3)] Two physical models of deep neural networks are proposed in this paper, one is pure ensemble deep neural network and the other is hybrid ensemble deep neural network. The former learning model corresponds to a quantum measurement of a microscopic state, such as CNN; the latter corresponds to a microscopically statistically averaged macroscopic state, such as RBM.

\item[(4)] The physical model of neurons in CNN is a quantum superposition of a quasi-particle incident wave (Figure \ref{figure1}) and is excited by the output. This excitation may be the elastic scattering caused by the incident wave (exit only includes the incident wave), or inelastic scattering (exit also includes internal new excited states), or various possible actions such as chemical reactions (exit only includes new quasiparticles). It obeys quantum mechanics. According to the superposition principle of quantum mechanics, the excitation output of a neuron is related not only to the intensity of incident quasi-particles in other neurons, but also to their coherence, and to their polarization direction or spin.

\item[(5)] The input of the deep learning network is treated as a wave function, and the image is also a wave. The state of the neuron is also expressed by a wave function. If the measured neuron is the number or probability of excited quasi-particles, the value of a layer of neurons in the neural network is a probability distribution. The deep learning operator should be performed on the complex number field, but because the activation function is ReLU, the computational difference in the real number domain may not be large.

\item[(6)]  Under such a physical model, the convolutional neural network algorithm is exactly the same as the quantum calculation method for measuring the number of quasi-particles excited by neurons, so that it can perfectly explain the technology of each important components of a convolutional neural network algorithm (convolution, rectification, activation, pooling, etc.) The purpose of convolution is to measure the number or probability of quasi-particles excited by each neuron. The convolution kernel is related to the Hamiltonian interaction potential of neurons. All neurons present an interference diffraction pattern - stripes and patches. That is, the deep convolutional neural network can measure the input wave vector or momentum, and its computational model can decompose white light into monochromatic light and decompose random-direction vibration into single-direction polarization. Therefore, the physical model of this paper can explain deep learning technology and success. This physical model shows that the deep convolutional neural network has natural learning ability and cognitive ability, and the model learns the ability to characterize the input micromechanical quantities, so it is reusable and can be applied across fields.

\item[(7)] The basis for parameter adjustment and optimization in the deep learning classification training process is the entropy in statistical physics, which is the number of microscopic states corresponding to the corresponding macroscopic state. Different types of training models should choose different cost functions according to the meaning of entropy, for example, the convolutional neural network model should use cross entropy as the objective function (cost function).

\item[(8)] A large number of operators, techniques, and methods in deep learning are related to the principles of physics such as energy, entropy, renormalization techniques, and translation operations; they are also related to physical world views such as symmetry, conjugacy, locality, hierarchy, etc.

\end{itemize}

The research in this paper shows that there are physics glimmers everywhere in deep learning. Deep learning techniques can be based on scientific theories. From the principles of physics, this dissertation presents the calculation methods for convolution calculation, pooling, normalization, and RBM, as well as the selection of cost functions, explains why deep learning must be deep, what characteristics are learned in deep learning, why convolutional neural networks do not have to be trained layer by layer, and the limitations of deep learning, etc. The physical model proposed in this paper can not only explain the successful technology of existing deep learning, but also predict many researchable directions and topics, such as positional neurons (these are in the research stage, and still need to be experimentally verified).

There is a striking homogeneity in the appearance and structure of the human cerebral cortex. In other words, the cerebral cortex uses the same calculations to accomplish all its functions. All the intelligence that humans exhibit (vision, auditory, physical movement...) are based on a unified set of algorithms. The deep learning technology is also based on a unified algorithm and is supported by physical theories. It will have a broad prospect for development.

%\section*{Acknowledgments}
%Helpful discussions with Tobias Osborne are gratefully acknowledged.
%This work was supported by the cluster of excellence EXC 201 ``Quantum Engineering and Space-Time Research''.

\nocite{*}
\bibliography{DL}{}
\bibliographystyle{unsrt}

\end{document}